\documentclass{article} % For LaTeX2e
\usepackage{iclr2027_conference,times}
\iclrfinalcopy

\usepackage{amsmath,amsfonts,bm}

\def\eqref#1{equation~\ref{#1}}
\def\1{\bm{1}}

\DeclareMathAlphabet{\mathsfit}{\encodingdefault}{\sfdefault}{m}{sl}
\SetMathAlphabet{\mathsfit}{bold}{\encodingdefault}{\sfdefault}{bx}{n}

\usepackage{hyperref}
\usepackage{url}
\usepackage{tikz}
\usepackage{booktabs}
\usepackage{tabularx}
\usepackage{multirow}
\usepackage{makecell}
\usepackage{graphicx}
\usepackage{placeins}
\usepackage{array}
\usepackage{xcolor}
\usepackage{colortbl}
\usepackage{enumitem}
\usepackage{fontawesome5}
\usepackage{hyperref}
\definecolor{oursbg}{RGB}{247,238,238}
\definecolor{oursrule}{RGB}{140,30,30}
\definecolor{groupbg}{RGB}{240,240,240}
\definecolor{stagebg}{RGB}{242,242,242}
\definecolor{statebg}{RGB}{244,244,244}
\definecolor{labelbg}{RGB}{244,244,244}
\definecolor{mutedgray}{RGB}{135,135,135}
\definecolor{baselinebg}{RGB}{255,243,243}
\definecolor{configbg}{RGB}{244,244,244}
\newcommand{\best}[1]{\textbf{#1}}
\newcommand{\second}[1]{\underline{#1}}
\newcommand{\NA}{\textcolor{gray!65}{--}}
\newcommand{\frameworkname}{SolveBind}

\title{Correct Answers, Closed Gates: \frameworkname{} for GUI Agents at the Verification Barrier}

\author{
Xinhao Song$^{1}$\thanks{Equal contribution.}, \quad
Haoxuan Peng$^{2}$\footnotemark[1], \quad
Su Su$^{3}$, \quad
Hongliang Wu$^{1}$, \quad
Sirui Song$^{1}$, \\
\textbf{Gongshen Liu}$^{1}$, \quad
\textbf{Linfeng Zhang}$^{1}$, \quad
\textbf{Dongrui Liu}$^{1}$\thanks{Corresponding authors.} \\
\\
$^{1}$Shanghai Jiao Tong University \quad
$^{2}$The University of Sydney \quad
$^{3}$Shandong University \\
\\
\textit{sxh001@sjtu.edu.cn} \quad \textit{haoxuan\_peng@outlook.com} \\
\\
\noindent
\textbf{Our code:}
\href{https://github.com/XinhaoS0101/SolveBind}{%
    \textcolor[HTML]{FF8000}{\faGithub\ SolveBind}%
}
\\
}

\begin{document}

\maketitle
% todo for sxh
\lhead{Preprint}
\begin{abstract}
Real-world websites and applications often deploy CAPTCHAs and other anti-automation mechanisms that use behavioral signals, such as interaction trajectories and timing, rather than outcomes alone, to distinguish human operators from automated agents. We refer to such settings as \emph{protected GUI environments}, where correctly solving a local verification task is insufficient for end-to-end task completion. An agent must also convert solver outputs into interactions accepted by the environment and adapt its execution based on runtime feedback under time constraints. For these environments, we introduce \frameworkname{}, a GUI-agent framework that combines \emph{hierarchical solving with runtime supervision}. A hierarchical solving tree decomposes verification into interdependent \emph{Perceive}, \emph{Ground}, \emph{Solve}, \emph{Act}, and \emph{Verify} capabilities. A runtime supervisor grounds action commitments in tool-derived evidence to allow them to be revised as new evidence arrives, and mediates execution under explicit recovery budgets. For verification, across nine CAPTCHA task families and nine backbone models, \frameworkname{} achieves the highest success rate on eight backbones, outperforming the strongest baseline by 13.3--34.5 percentage points. For end-to-end task completion, it achieves the highest success rate on all six backbones evaluated on AndroidWorld. In a paired GPT-5.5 analysis, MobileAgent produces correct answers but fails dynamic verification on 13 of 90 tasks; \frameworkname{} succeeds on seven of these tasks. 
With automatic submission retained, removing binding reduces success from 82.22\% to 71.56\% and increases latency and token usage. Recovery caps reduce execution cost with a small observed success trade-off: removing them raises success from 82.22\% to 83.56\%, while increasing latency and token usage to 3.7$\times$ and
2.4$\times$ their full-supervision values, respectively. These results support evidence-grounded execution control while identifying a success--cost trade-off in recovery and failure boundaries of heuristic submission rules.
\end{abstract}

\section{Introduction}
\label{sec:introduction}

GUI agents aim to translate users' task intents into complete interaction workflows, transforming requests such as retrieving information, filling out forms, or managing files into action sequences that span multiple pages and applications~\citep{webvoyager2024,agents2025,androidworld2024}. Real-world services, however, often deploy anti-automation mechanisms that rely on behavioral signals, such as interaction trajectories and timing, rather than outcomes alone, to distinguish human operators from automated agents~\citep{becaptchamouse2022,fpagent2026}. Even if an agent has identified a viable solution, it must still advance the task through interactions accepted by the environment within the available time budget, so verification checkpoints become an integral part of task execution rather than an external obstacle to it.

\begin{figure}[t]
    \vspace{-6pt}
    \centering
    \includegraphics[width=0.9\linewidth]{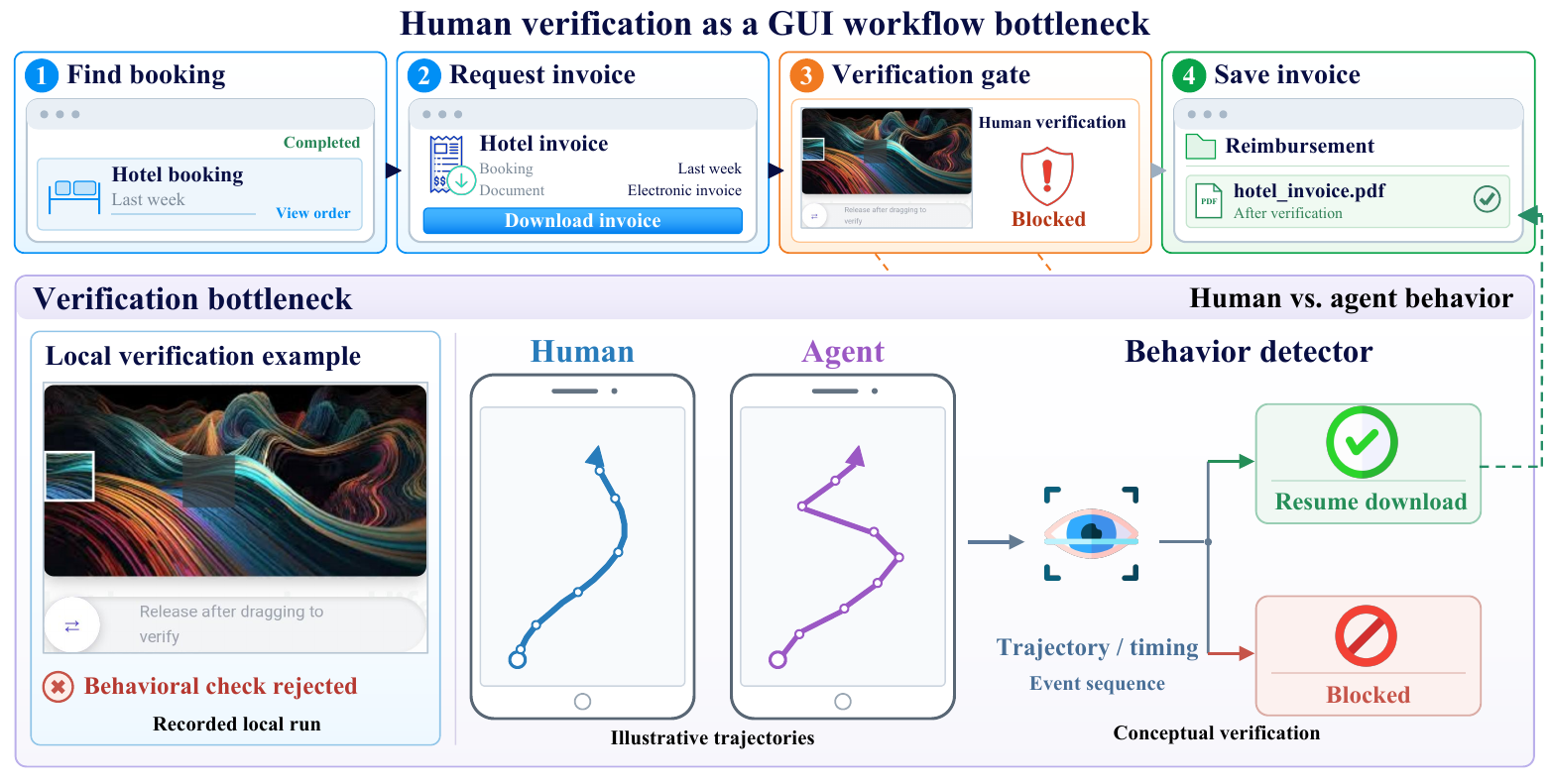}
    \caption{Human verification as a bottleneck in a GUI workflow. \textbf{Top:} a verification gate blocks an invoice-download task. \textbf{Bottom left:} a correct slider position is still rejected. \textbf{Bottom right:} the detector scores trajectory and timing, so human-unlike motion is blocked even at the correct target.}
    \label{fig:motivation}
    \vspace{-10pt}
\end{figure}
Consider the invoice-download task in Figure~\ref{fig:motivation}. The agent locates the booking, requests the invoice, and reaches a human-verification gate before saving the file. Although it correctly identifies the target, the attempt can still be rejected because the verification mechanism also evaluates behavioral signals such as the drag trajectory and timing~\citep{becaptchamouse2022,hll2026}. The bottleneck therefore lies not only in solving the local challenge, but also in executing the solution in a manner accepted by the environment. Re-solving may not resolve the rejection, while repeated attempts consume the limited verification budget and delay task completion. This illustrates why protected GUI environments require agents to coordinate local solving with feedback-aware execution. 

Interactive verification and behavioral discrimination make this a broader execution problem rather than an issue confined to a single CAPTCHA type. HLL\citep{hll2026} jointly evaluates task responses and interaction evidence in dynamic CAPTCHA tasks, revealing a gap between correctly solving a challenge and passing verification. Turing Test on Screen\citep{turing_screen2026} further shows that touch behavior alone can distinguish agents from humans. We refer to GUI environments equipped with anti-automation mechanisms that monitor interaction behavior and distinguish automated from human operation as \emph{protected GUI environments}. Such evidence may derive from a single action, a local interaction sequence, or a longer execution history, shaping the conditions under which subsequent execution proceeds~\citep{becaptchamouse2022,hll2026,fpagent2026}. We focus on two outcomes in these environments: task success rate and end-to-end latency.

Existing GUI-agent frameworks combine hierarchical planning with specialized tools to improve long-horizon decision making and interface grounding~\citep{agents2025,agents22025}. In protected settings, however, these capabilities must be coordinated with the agent's current execution state. Redundant actions and costly planning or reflection calls increase end-to-end latency~\citep{osworld_human2025}, while undetected action failures can lead to ineffective repetition and error accumulation~\citep{verigui2026}. The central challenge is therefore determining \emph{which solver outputs should constrain execution and which observations should cause the agent to revise its next action}. Our key idea is to treat solver outputs as runtime evidence that can be retained and revised. When the evidence supports a concrete action, the runtime records it as an \emph{action commitment} and prioritizes that action until new observations update that evidence or trigger reconsideration. This coordination is motivated by rational metareasoning: computation should be weighed against both its cost and its expected value for the next decision~\citep{russell1991principles, hay2012selecting}.

Building on this idea, we introduce \frameworkname{}, a GUI-agent framework combining hierarchical tool-based solving with runtime supervision. A hierarchical solving tree organizes capability dependencies and alternative tool branches, allowing the model to select the capabilities needed for the current execution state. The runtime supervisor binds tool-derived evidence to action commitments and, under full supervision, prioritizes admissible pending commitments over conflicting action proposals. Execution feedback then updates a shared protocol state that governs workflow progression, verification, and bounded recovery, enabling subsequent attempts to preserve still-valid evidence and revise affected assumptions invalidated by new observations. Task adapters connect this protocol to different environments through task-specific primitives, tools, and completion interfaces.

We evaluate \frameworkname{} on nine HLL dynamic verification families~\citep{hll2026} and AndroidWorld~\citep{androidworld2024}, across nine foundation models and three baseline harnesses. On verification tasks, \frameworkname{} achieves the highest success rate on eight of nine backbones, exceeding the strongest baseline by 13.3--34.5 percentage points. On general GUI tasks, it leads on all six evaluated backbones while reducing end-to-end latency in most comparisons. A paired analysis finds 13 correct-answer dynamic rejections under MobileAgent across 90 GPT-5.5 tasks, seven of which \frameworkname{} resolves. With automatic submission fixed, relaxing binding reduces success from 185/225 to 161/225, isolating the effect of dispatch precedence. 
% Removing only the recovery caps raises success by 1.33 percentage points but increases latency and token usage to 3.7$\times$ and 2.4$\times$ the full-supervision values.

\begin{itemize}[leftmargin=*,itemsep=1pt,topsep=2pt,parsep=0pt]
    \item We formulate GUI task execution in \emph{protected GUI environments}, distinguishing goal achievement from environmental acceptance.
    \item We formulate three design principles from rational metareasoning: evidence-grounded commitment, progress-checked transition, and feedback-directed bounded recovery. We instantiate these three principles in \frameworkname{}, which pairs a hierarchical solving tree with a runtime supervisor that binds tool evidence to revisable action commitments.
    \item We evaluate \frameworkname{} across nine backbones, three baseline harnesses, and two environments, showing success and latency improvements in multiple configurations and using ablations to assess the effects of runtime-control configurations.
\end{itemize}
\vspace{-4pt}
\section{Related Work}
\label{sec:related_work}
\vspace{-3pt}

\textbf{GUI agent harnesses and native policies.}
GUI-agent research follows two complementary directions. External harnesses organize multimodal models through planning, tool use, and memory: WebVoyager~\citep{webvoyager2024} supports web interaction; MobileAgent~\citep{wang2024mobile} and AppAgent~\citep{appagent2024} coordinate mobile perception and action; Agent S~\citep{agents2025} uses experience-augmented hierarchical planning; and Agent S2~\citep{agents22025} combines a generalist planner with specialist grounding. Native policies, by contrast, directly learn interface grounding~\citep{seeclick2024,uground2025,cogagent2024} or integrate perception, reasoning, and action generation~\citep{uitars2025}. Evaluation spans action prediction from recorded interaction traces~\citep{mind2web2023,androidinthewild2023} and task completion in executable environments~\citep{webarena2024,osworld2024,androidworld2024}. These settings primarily assess grounding accuracy and goal achievement, rather than whether execution is accepted based on interaction behavior. \frameworkname{} builds on the external-harness paradigm but explicitly coordinates task solving with environmental acceptance: an action may target the correct interface element yet still be rejected, requiring execution to account for both the intended effect and the conditions under which the environment permits progress.

\textbf{Execution verification and recovery.}
Existing frameworks differ in where verification intervenes, what execution decisions it constrains, and how recovery is bounded. VeriSafe~\citep{verisafe2025} translates user instructions into logical specifications and checks proposed actions before execution, advising on predicate mismatches while blocking critical violations. VLAA-GUI~\citep{vlaa_gui2026} verifies action effects and completion claims, using repetition detection and strategy-switching directives to constrain subsequent decisions while leaving replacement actions to its manager. StructAgent~\citep{structagent2026} combines low-level effect checks with verifier-backed milestone updates and can return execution to the planner. MAPLE~\citep{maple2025} compares observed states with predicted postconditions and constructs recovery plans, returning to high-level replanning after two recovery failures. MobileUse~\citep{li2025mobileuse} uses hierarchical, on-demand reflection to support recovery while limiting overhead. VeriGUI~\citep{verigui2026} learns effect prediction and self-correction within the policy, rather than relying on an independent dispatch controller. In \frameworkname{}, every action dispatch passes through the supervisor. Under full control, admissible pending commitments take precedence over conflicting action proposals. Execution feedback updates their supporting evidence, preserving valid commitments while triggering reconsideration when that support no longer holds. Recovery is budgeted by category, with attempts and refreshes tracked separately. Our focus is therefore not verification alone, but the coordination of evidence-supported dispatch with rejection-aware, bounded recovery. 
More broadly, ReAct and Reflexion incorporate environmental or evaluative feedback into reasoning~\citep{react2023,reflexion2023}, whereas intrinsic self-correction may fail without reliable external feedback~\citep{huang2024cannot}. Our baseline cases, detailed in Appendix~\ref{app:baseline_cases}, reveal a complementary failure mode: an agent may correctly diagnose a failure yet still repeat the rejected action.

\textbf{Protected interaction and execution efficiency.}
CAPTCHA research has documented vulnerabilities in semantic image challenges~\citep{sivakorn2016robot} and the human burden of deployed verification~\citep{searles2023captcha}. Recent solvers exploit language and vision-language models: Halligan formulates visual CAPTCHA solving as search~\citep{teoh2025halligan}, while Oedipus decomposes reasoning CAPTCHAs into DSL-guided substeps~\citep{deng2025oedipus}. Salman and Bicakci~\citep{salman2026capagenticera} study adaptation across CAPTCHA encounters by using VLM outputs to update a detector; our focus is evidence reuse and revision within task execution. Open CaptchaWorld evaluates multimodal agents on diverse interactive CAPTCHA puzzles~\citep{luo2025opencap}. Next-Gen CAPTCHAs develops scalable interactive challenges around persistent human--agent capability gaps~\citep{liu2026next}.
Beyond puzzle solving, behavioral detection considers how interactions are performed: BeCAPTCHA-Mouse~\citep{becaptchamouse2022} distinguishes human and synthetic input using trajectory-level neuromotor features, while FP-Agent~\citep{fpagent2026} finds behavioral fingerprints more discriminative than shared browser fingerprints. HLL~\citep{hll2026} evaluates answers and interaction evidence, while Turing Test on Screen~\citep{turing_screen2026} measures the distinguishability of agent-generated touch behavior. Together, these studies motivate separating local challenge solving from environmental acceptance. ReCAP~\citep{recap2026} addresses CAPTCHA solving through trajectory generation and self-corrective fine-tuning of native GUI policies. Its interactive benchmark evaluates puzzle solving but does not simulate behavioral biometrics such as cursor speed or acceleration. In contrast, \frameworkname{} leaves the backbone unchanged and coordinates execution at runtime under interaction-dependent validation. The distinction concerns the acceptance criterion and execution-control mechanism, rather than whether the task itself is interactive. Finally, redundant actions and expensive planning or reflection increase end-to-end latency~\citep{osworld_human2025}, motivating joint evaluation of task success and completion time rather than treating rapid failure as efficient.

\textbf{Runtime supervision and metareasoning.}
The separation of action proposals from external control is well established in supervisory control of discrete-event systems~\citep{ramadge_wonham1987} and shielded reinforcement learning~\citep{alshiekh2018shielding}. We adopt this architectural separation, but not its formal guarantees: admissibility is determined by runtime predicates over accumulated evidence, without guarantees of correctness or non-blocking behavior. Our design rationale instead draws on metareasoning, which weighs computation against its cost and expected value for subsequent decisions~\citep{horvitz1987reasoning, russell1991principles,zilberstein1996anytime, hay2012selecting}, as well as resource-rational accounts of cognition~\citep{lieder2020resource}. These perspectives motivate acting once sufficient evidence is available, revising commitments when their supporting evidence changes, and bounding recovery once further computation is unlikely to improve subsequent decisions. These principles remain empirical design choices rather than consequences of an optimal metareasoning policy. In our ablations (Section~\ref{sec:runtime_ablation}), relaxing mediation, fail-safe defaults, or supervision reduces success rates while increasing latency and token usage. Removing only the recovery cap yields slightly higher observed success at substantially greater latency and token usage, supporting bounded recovery as a practical success-cost trade-off.
\section{Problem Formulation and Design Rationale}
\label{sec:preliminaries}

\subsection{GUI task execution in protected environments}
\label{sec:protected_gui_tasks}
\label{sec:runtime_problem}

We study GUI task execution on websites and applications equipped with human-verification or anti-automation mechanisms. A \textbf{protected GUI environment} is one in which these mechanisms affect whether interactions are accepted and whether the workflow can proceed. We represent such an environment as $\mathcal{E}^{\mathrm{prot}}=(\mathcal{E},\mathcal{V})$ with $\mathcal{V}=\{(\phi_j,\mathcal{V}_j)\}_{j=1}^{K}$, where $\mathcal{E}$ is the GUI environment, $\phi_j$ extracts execution evidence, and $\mathcal{V}_j$ evaluates that evidence in context. Evidence may concern a single action, a local interaction sequence, or a longer history. The mechanisms may use it to assess interaction validity or distinguish human from automated behavior.

\begin{figure}[t]
    \vspace{-6pt}
    \centering
    \includegraphics[width=0.95\linewidth]{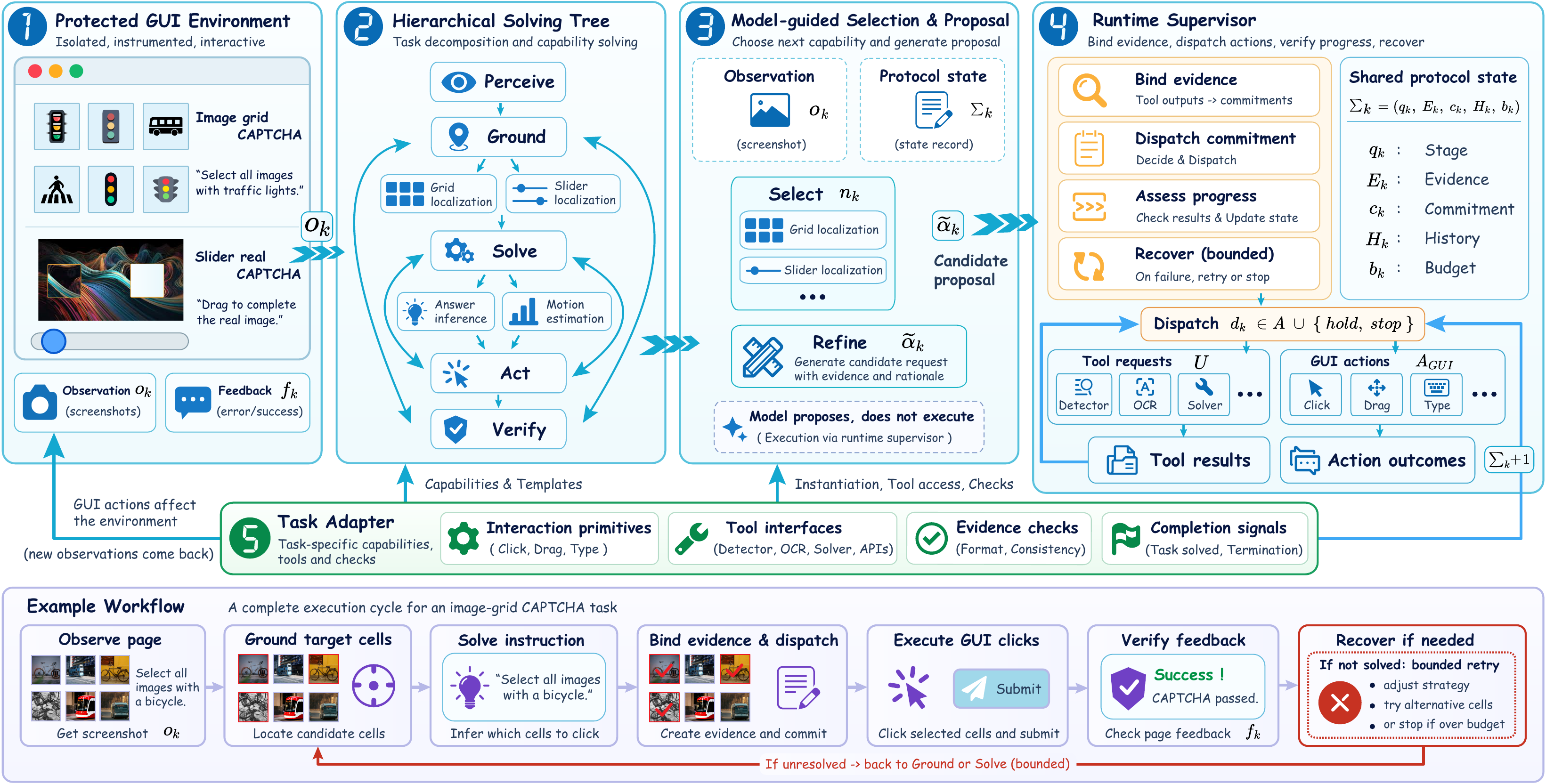}
    \caption{Overview of \frameworkname{}. \textbf{(1)}--\textbf{(5)} illustrate the protected GUI environment, hierarchical solving tree, model, runtime supervisor, and task adapter. The runtime supervisor mediates execution so that only dispatched operations are passed to external tools or the GUI. The bottom panel shows a representative execution cycle for category-guided image selection.}
    \label{fig:method_overview}
    \vspace{-12pt}
\end{figure}

Given a user goal $g$, the agent receives observations $o_t$ and executes GUI actions $a_t\in\mathcal{A}_{\mathrm{GUI}}$. Actions generate execution evidence, such as interaction events and timing, that the protection mechanisms may evaluate. These evaluations can permit continued execution, require renewed verification, or block progress. They therefore form part of the environment dynamics rather than serving only as a terminal check. The agent has access only to observations and feedback exposed by the environment, not to its internal states or undisclosed verification rules.

Let $\tau^{\mathrm{prot}}$ denote a terminated execution record. We distinguish goal achievement, $\mathrm{Goal}_g$, from satisfaction of the verification requirements for completion, $\mathrm{Accept}_{g,\mathcal{V}}$. Task success requires both:
\begin{equation}
    S_g\big(\tau^{\mathrm{prot}}\big)
    =\mathbf{1}\left[
        \mathrm{Goal}_g\big(\tau^{\mathrm{prot}}\big)
        \land\mathrm{Accept}_{g,\mathcal{V}}\big(\tau^{\mathrm{prot}}\big)
    \right].
    \label{eq:protected_task_success}
\end{equation}

We assess success jointly with end-to-end wall-clock time,
\begin{equation}
    L\big(\tau^{\mathrm{prot}}\big)=t_{\mathrm{end}}-t_{\mathrm{start}},
    \label{eq:end_to_end_latency}
\end{equation}
measured from the first model call until success or failure termination. This interval includes model inference, tool execution, environmental responses, verification, and recovery. We report success jointly with mean latency over all evaluated runs, including failures. A lower all-run mean does not by itself establish faster successful completion, since methods may terminate unsuccessfully or succeed on different task subsets. Appendix~\ref{app:problem_formulation} formalizes the trajectories, verification rules, acceptance conditions, and timing comparisons.

\subsection{Design principles from a rational metareasoning perspective}
\label{sec:design_rationale}

Protected GUI execution requires deciding whether to continue solving, act on available evidence, or reassess an attempt. Additional computation and evidence gathering may improve the next decision, but both consume execution time. Rational metareasoning weighs the expected decision benefits of computation against its costs~\citep{russell1991principles,hay2012selecting}. We use this perspective to guide three principles for coordinating computation and execution.

\textbf{Evidence-grounded commitment.} Maintain an evidence-supported operation as a revisable commitment. While its supporting conditions hold and execution constraints are satisfied, let the commitment constrain conflicting proposals and redundant solving. Reconsider it when new observations indicate that relevant conditions have changed.

\textbf{Progress-checked transition.} Advance only when available evidence satisfies the completion criterion for the current capability or the overall task, not merely because an operation has been dispatched. When an outcome remains unresolved, seek evidence that helps determine whether to advance, reassess, or continue observing.

\textbf{Feedback-directed bounded recovery.} Use feedback and execution history to revise affected hypotheses and commitments while preserving applicable evidence. Focus recovery on unresolved issues and bound the corresponding recovery operations with explicit budgets. Appendix~\ref{app:metareasoning_rationale} develops the rationale and scope of these principles. 
\section{\frameworkname{}: Hierarchical Solving with Runtime Supervision}
\label{sec:method}

\frameworkname{} uses a hierarchical solving tree to organize task-solving capabilities and a runtime supervisor to coordinate execution, progress verification, and recovery. Tool-derived evidence supports concrete, revisable action commitments, while execution feedback updates a shared protocol state that guides subsequent capability selection and dispatch.

\subsection{Framework overview}
\label{sec:framework_overview}
\label{sec:framework_formulation}
As shown in Figure~\ref{fig:method_overview}, the solving tree specifies required subgoals and alternative capabilities. The model selects capabilities and proposes operations, while the supervisor determines which operations to dispatch based on retained evidence, pending commitments, and execution constraints. Execution feedback then informs subsequent decisions. 

At runtime decision step $k$, the framework maintains the shared protocol state
\begin{equation}
    \Sigma_k=(q_k,E_k,c_k,\mathcal{H}_k,b_k),
    \label{eq:protocol_state}
\end{equation}
where $q_k$ is the active stage, $E_k$ is the retained evidence, $c_k$ is a pending commitment or $\emptyset$, $\mathcal{H}_k$ records the outcome history, including failed hypotheses, and $b_k$ contains the remaining recovery budgets. This state provides a common context for capability selection, dispatch, and recovery.

The model generates a proposal $\tilde{a}_k$ from the current observation and its view $\Sigma_k^\pi$ of the protocol state. The operation space is $\mathcal{A}=\mathcal{U}\cup\mathcal{A}_{\mathrm{GUI}}$, comprising tool requests $\mathcal{U}$ and direct GUI actions $\mathcal{A}_{\mathrm{GUI}}$. The supervisor returns $d_k\in\mathcal{A}\cup\{\mathrm{hold},\mathrm{stop}\}$. When $d_k\in\mathcal{A}$, the runtime dispatches the selected operation; $\mathrm{hold}$ defers dispatch, and $\mathrm{stop}$ terminates the run. After each nonterminal decision, the runtime uses available observations and feedback to update the protocol state to $\Sigma_{k+1}$, which informs the next capability selection and dispatch. Appendix~\ref{app:protocol_cycle} specifies the complete interfaces and execution cycle.

\begin{table}[t]
    \vspace{-6pt}
    \centering
    \caption{Dynamic-verification performance across backbones and agent policies. Each policy reports SR (\%, higher is better), E2E (s, lower is better), and Tok (k, lower is better). \textbf{Bold} and \underline{underline} mark the best and second-best SR per backbone; ``--'' denotes unavailable results.}
    \label{tab:hll_main}
    \scriptsize
    \setlength{\tabcolsep}{3.2pt}
    \renewcommand{\arraystretch}{1.15}
    \begin{tabular}{@{}l*{3}{>{\columncolor{oursbg}}r}*{9}{r}@{}}
        \toprule
        & \multicolumn{3}{c}{\textbf{\frameworkname{}}} & \multicolumn{3}{c}{\textbf{MobileAgent}} & \multicolumn{3}{c}{\makecell{\textbf{UI-TARS}\\\textbf{Harness}}} & \multicolumn{3}{c}{\textbf{AppAgent}} \\
        \cmidrule[1.3pt](lr){2-4}
        \cmidrule(lr){5-7}
        \cmidrule(lr){8-10}
        \cmidrule(l){11-13}
        \textbf{Backbone} & \cellcolor{oursbg}\textbf{SR} & \cellcolor{oursbg}\textbf{E2E} & \cellcolor{oursbg}\textbf{Tok} & \textbf{SR} & \textbf{E2E} & \textbf{Tok} & \textbf{SR} & \textbf{E2E} & \textbf{Tok} & \textbf{SR} & \textbf{E2E} & \textbf{Tok} \\ 
        \midrule
        \rowcolor{groupbg}
        \multicolumn{13}{@{}l}{\textbf{Policy backbones}} \\
        GPT-5.5 & \textbf{80.0} & 112.0 & 37.80 & \underline{66.7} & 167.3 & 23.82 & 43.3 & 189.3 & 59.30 & 26.7 & 207.8 & 33.66 \\
        GPT-5.4 & \textbf{46.7} & 40.6 & 74.93 & 11.1 & 162.9 & 37.37 & \underline{15.6} & 91.5 & 73.01 & 6.7 & 195.1 & 56.34 \\
        Claude Opus 4.8 & \textbf{75.6} & 85.4 & 77.54 & \underline{61.1} & 208.9 & 52.72 & 23.3 & 183.7 & 81.54 & 24.4 & 211.2 & 81.66 \\
        Claude Opus 4.6 & \underline{34.4} & 185.2 & 47.76 & \textbf{38.9} & 202.2 & 30.75 & 10.0 & 265.7 & 77.38 & 3.3 & 230.6 & 23.43 \\
        Gemini-3.1-Pro & \textbf{76.7} & 88.6 & 59.44 & \underline{42.2} & 577.7 & 63.97 & 15.6 & 331.4 & 92.02 & 25.6 & 248.4 & 26.27 \\
        GLM-5v-turbo & \textbf{57.8} & 140.4 & 97.49 & \underline{38.9} & 262.8 & 42.80 & 12.2 & 114.1 & 77.92 & 15.6 & 154.4 & 38.37 \\
        MiniMax-M2.7 & \textbf{32.2} & 152.3 & 68.73 & 0.0 & 316.2 & 28.29 & \underline{10.0} & 76.2 & 22.12 & 0.0 & 87.6 & 8.97 \\
        MiniMax-M3 & \textbf{43.3} & 261.2 & 135.54 & 6.7 & 555.8 & 76.08 & \underline{12.2} & 220.2 & 80.48 & 11.1 & 195.2 & 43.46 \\
        Qwen3.7-plus & \textbf{65.6} & 126.3 & 56.99 & \underline{44.4} & 621.1 & 67.49 & 28.9 & 300.9 & 84.42 & 21.1 & 264.6 & 31.93 \\
        \addlinespace[1.5pt]
        \rowcolor{groupbg}
        \multicolumn{13}{@{}l}{\textbf{Specialized GUI agent}} \\
        UI-TARS-1.5-7B ($n=900$) & \NA & \NA & \NA & \NA & \NA & \NA & 36.11 & 116.38 & 19.91 & \NA & \NA & \NA \\
        \bottomrule
    \end{tabular}
    \vspace{-8pt}
\end{table}
\FloatBarrier

\subsection{Hierarchical capability organization}
\label{sec:capability_organization}

The solving tree uses a predefined skeleton of \emph{Perceive}, \emph{Ground}, \emph{Solve}, \emph{Act}, and \emph{Verify} capabilities for task perception, target grounding, solution generation, interaction execution, and outcome verification, respectively. The task adapter instantiates applicable branches according to the task's interaction requirements and available tools, linking tool selection to the current solving needs and evidence dependencies. CAPTCHA solving provides the primary instantiation of this structure. Formally,
\begin{equation}
    \begin{aligned}
        \mathcal{T}_g
        =(\mathcal{N}_g,E_{\land},E_{\lor},\mathrm{Spec}),\ \ \ 
        \mathrm{Spec}(n)
        =(\mathrm{Pre}_n,\mathrm{Act}_n,\mathrm{Done}_n).
    \end{aligned}
    \label{eq:solving_tree}
\end{equation}
Here, $\mathcal{N}_g$ is the set of capability nodes, $E_{\land}$ encodes required subgoals, and $E_{\lor}$ encodes alternative capabilities. Each node $n$ is specified by an entry predicate $\mathrm{Pre}_n(\Sigma)$, a set of parameterized operations $\mathrm{Act}_n$, and a completion predicate $\mathrm{Done}_n(\Sigma)$. 
% The entry and completion predicates capture evidence dependencies across capabilities: grounding evidence supports subsequent interaction, while evidence satisfying completion criteria permits advancement.

For example, grounding in grid and slider tasks uses different target descriptions, geometric information, and tools. A single tool may also provide both grounding evidence and a candidate solution. Appendix~\ref{app:capability_details} further describes the node and dependency semantics.

\subsection{Runtime supervision}
\label{sec:runtime_contract}

\textbf{Evidence binding and commitment-based dispatch.}
\label{sec:action_commitment}
When a tool output passes the binding checks, the runtime establishes a commitment $c_k=(n,a^c,E^c)$, linking a concrete operation $a^c$ to the capability node $n$ which it serves and its supporting evidence $E^c\subseteq E_k$. In the full-control configuration, an admissible pending commitment takes dispatch precedence:
\begin{equation}
    \begin{aligned}
        c_k\ne\emptyset\ \land\
        \mathrm{Admissible}(c_k,\Sigma_k)
        \qquad\Longrightarrow\quad d_k=a^c.
    \end{aligned}
    \label{eq:commitment_precedence}
\end{equation}
Admissibility requires that the supporting evidence remains valid in the current context, the relevant stage and budget checks pass, and no terminal condition has been triggered. If a proposal conflicts with an admissible commitment, the supervisor replaces it with the committed operation $a^c$.

Once dispatched, a commitment is no longer pending for repeated execution; its supporting evidence and attempt record are retained for outcome assessment. Dispatch alone does not establish progress or task completion.

For example, the supervisor can replace a repeated localization proposal with an admissible pending operation already supported by grounding evidence. The commitment remains subject to reassessment if the target or page conditions change. Appendix~\ref{app:evidence_details} details the binding checks and commitment lifecycle.

\textbf{Progress verification and bounded recovery.}
\label{sec:progress_recovery}
The supervisor assesses outcomes as $\mathrm{progress}$, $\mathrm{failure}$, or $\mathrm{unresolved}$ using the active stage, execution feedback, and subsequent observations. Progress evidence updates capability status, with advancement permitted once the corresponding completion criterion is satisfied. Failure feedback guides revisions to hypotheses and commitments or the selection of an alternative branch, while unresolved outcomes prompt further observation and feedback acquisition. 
% These assessments update the shared protocol state so that subsequent capability selection and dispatch reflect the latest execution results.

Recovery combines runtime constraints with model guidance, using feedback and attempt history to revise affected hypotheses while preserving applicable evidence. Recovery budgets are consumed in configured units, such as attempts or refreshes. Once a budget is exhausted, the runtime stops the recovery operations within its scope and terminates the run when no permitted continuation remains. Appendix~\ref{app:recovery_details} specifies the feedback categories, state updates, and budget rules.

\vspace{-2pt}
\section{Experiments}
\label{sec:experiments}
\vspace{-2pt}
We evaluate \frameworkname{} for reliable and efficient task execution in protected GUI environments from two complementary perspectives: interaction acceptance on the verification side and workflow efficiency on the task side. Appendices~\ref{app:repeated_run_results}, \ref{app:baseline_cases}, and~\ref{app:solvebind_cases} provide statistical stability analysis, failure-mode analysis, and case studies, respectively.

\vspace{-2pt}

\subsection{Experimental setup}
\label{sec:experimental_setup}
\vspace{-2pt}

\textbf{Evaluation settings.} On the verification side, we evaluate CAPTCHA instances from nine HLL task families with \emph{dynamic interaction validation}~\citep{hll2026}, using distraction level~2 and \texttt{hard} difficulty where supported, otherwise \texttt{normal}. Success requires both a correct answer and the corresponding interaction evidence. On the task side, we use AndroidWorld~\citep{androidworld2024} to assess workflow completion and execution efficiency. Appendices~\ref{app:env_init} and~\ref{app:budgets_termination} document the environments, episode initialization, and execution limits.

\textbf{Evaluated approaches.} We compare \frameworkname{} with the framework-driven MobileAgent~\citep{wang2024mobile} and AppAgent~\citep{appagent2024} across foundation models. These harnesses organize execution through planning, tools, memory, and feedback. We also evaluate the released UI-TARS-1.5-7B checkpoint with its native harness~\citep{uitars2025, bytedance2025uitars15}. We additionally adapt the UI-TARS harness to the same general-purpose foundation models and denote this configuration \emph{UI-TARS Harness}. Appendices~\ref{app:agent_integration} and~\ref{app:interface_boundaries} describe agent integration and information-access boundaries.

\textbf{Backbone models.} We evaluate nine foundation-model backbones on HLL: GPT-5.5 and GPT-5.4~\citep{openai2026gpt55,openai2026gpt54}, Claude Opus 4.8 and 4.6~\citep{anthropic2026opus48,anthropic2026opus46}, Gemini 3.1 Pro Preview~\citep{google2026gemini31pro}, GLM-5V-Turbo~\citep{glm2026glm5vturbo}, MiniMax-M2.7 and MiniMax-M3~\citep{minimax2026m27,minimax2026m3}, and Qwen3.7-Plus~\citep{qwen2026qwen37}. The AndroidWorld evaluation uses six of these backbones: GPT-5.5, Claude Opus 4.8, Gemini 3.1 Pro Preview, GLM-5V-Turbo, MiniMax-M3, and Qwen3.7-Plus. UI-TARS-1.5-7B is evaluated separately as a specialized GUI policy rather than counted among these nine backbones.

\textbf{Metrics.} We report task success rate (SR), end-to-end wall-clock time (E2E), and mean total-token usage (Tok). E2E starts with the first model call and ends at success or failure termination, including model inference, tool calls, GUI operations, environmental responses, and recovery. Tok sums the input and output tokens from all model calls within each task run and averages these totals across runs. Appendix~\ref{app:metric_aggregation} specifies the scored populations and aggregation rules.

\vspace{-3pt}

\subsection{Results on verification-side interaction acceptance}
\label{sec:hll_results}
\vspace{-3pt}

\textbf{Coordinating solving and execution improves interaction acceptance.} On the verification side, success requires executing a solution in a form accepted by the environment. In Table~\ref{tab:hll_main}, \frameworkname{} achieves the highest success rate on eight of nine backbones, exceeding the strongest baseline by 13.3--34.5 percentage points across those eight configurations. Appendix~\ref{app:full_quantitative_analysis} provides per-backbone success gaps and unscaled token results.

\begin{table}[t]
    \vspace{-6pt}
    \centering
    \caption{AndroidWorld performance across backbones and agent policies. Each policy reports SR (\%, higher is better), E2E (s, lower is better), and Tok (k, lower is better). \textbf{Bold} and \underline{underline} mark the best and second-best SR per backbone.}
    \label{tab:androidworld_main}
    \scriptsize
    \setlength{\tabcolsep}{3.2pt}
    \renewcommand{\arraystretch}{1.15}
    \begin{tabular}{@{}l*{3}{>{\columncolor{oursbg}}r}*{9}{r}@{}}
        \toprule
        & \multicolumn{3}{c}{\textbf{\frameworkname{}}} & \multicolumn{3}{c}{\textbf{MobileAgent}} & \multicolumn{3}{c}{\makecell{\textbf{UI-TARS}\\\textbf{Harness}}} & \multicolumn{3}{c}{\textbf{AppAgent}} \\
        \cmidrule[1.3pt](lr){2-4}
        \cmidrule(lr){5-7}
        \cmidrule(lr){8-10}
        \cmidrule(l){11-13}
        \textbf{Backbone} & \cellcolor{oursbg}\textbf{SR} & \cellcolor{oursbg}\textbf{E2E} & \cellcolor{oursbg}\textbf{Tok} & \textbf{SR} & \textbf{E2E} & \textbf{Tok} & \textbf{SR} & \textbf{E2E} & \textbf{Tok} & \textbf{SR} & \textbf{E2E} & \textbf{Tok} \\
        \midrule
        \rowcolor{groupbg}
        \multicolumn{13}{@{}l}{\textbf{General foundation models}} \\
        GPT-5.5 & \best{72.4} & 196.88 & 100.86 & \second{40.52} & 838.03 & 173.47 & 37.07 & 179.54 & 29.38 & 35.3 & 282.2 & 45.74 \\
        Claude Opus 4.8 & \best{62.1} & 189.43 & 135.59 & \second{40.52} & 806.0 & 256.18 & 34.48 & 475.13 & 44.58 & 26.7 & 159.1 & 29.17 \\
        Gemini 3.1 Pro & \best{61.2} & 272.96 & 83.67 & \second{31.03} & 1008.0 & 160.94 & 27.59 & 332.37 & 23.54 & 25.9 & 282.3 & 29.72 \\
        GLM-5v-turbo & \best{54.3} & 274.27 & 134.37 & 6.03 & 1372.2 & 310.90 & 15.52 & 289.05 & 63.44 & \second{25.0} & 349.8 & 53.48 \\
        MiniMax-M3 & \best{39.7} & 211.74 & 136.87 & 24.14 & 671.62 & 269.49 & \second{31.03} & 163.59 & 30.97 & 16.4 & 268.9 & 15.14 \\
        Qwen3.7-plus & \best{49.1} & 368.84 & 92.59 & \second{25.00} & 1187.57 & 228.95 & 16.38 & 231.52 & 31.35 & 19.8 & 318.4 & 40.16 \\
        \addlinespace[1.5pt]
        \rowcolor{groupbg}
        \multicolumn{13}{@{}l}{\textbf{Specialized GUI agent}} \\
        UI-TARS-1.5-7B & \NA & \NA & \NA & \NA & \NA & \NA & 19.83 & 95.24 & 14.02 & \NA & \NA & \NA \\
        \bottomrule
    \end{tabular}
    \vspace{-10pt}
\end{table}
\FloatBarrier

\textbf{Higher verification success with lower latency.} Relative to MobileAgent, GPT-5.5 and Gemini 3.1 Pro Preview improve SR while reducing mean E2E. GPT-5.5 reduces E2E from 167.3 to 112.0 seconds, a 33.05\% reduction. Gemini 3.1 Pro Preview reduces E2E from 577.7 to 88.6 seconds, an 84.66\% reduction, while increasing success by 34.5 percentage points based on the displayed SR values. Token usage increases for GPT-5.5 but decreases for Gemini 3.1 Pro Preview and Qwen3.7-Plus. Lower latency is therefore not uniformly accompanied by lower token usage.

\textbf{GUI specialization alone does not remove the verification bottleneck.} UI-TARS-1.5-7B with its native harness completes 325 of 900 tasks in the full evaluation, achieving 36.11\% success and leaving 575 tasks unsuccessful. Thus, a policy specialized for GUI operation still encounters a substantial completion bottleneck under difficult, high-distraction verification conditions. Solving, interaction evidence, and feedback coordination remain central to translating GUI capabilities into successful protected execution. Appendix~\ref{app:ui_tars_native_results} and Appendix~\ref{app:mechanism_trace_statistics} report the native-policy evaluation separately from the backbone--harness comparisons.
% and report how often the supervisor changes dispatch decisions, using step-level logs from a separate batch.

\textbf{Correct answers can still fail dynamic verification.} A paired diagnostic analysis of 90 HLL tasks with GPT-5.5 identifies 13 MobileAgent episodes (14.44\%) whose submitted answers pass the static correctness check but are rejected by the dynamic validator and end in failure. \frameworkname{} succeeds on seven of these same tasks (7/13, 53.85\%). Across the full diagnostic panel, 13 tasks change from MobileAgent failure to \frameworkname{} success, while one changes in the opposite direction, a net gain of 12/90 or 13.33 percentage points. This establishes the presence of the targeted failure mode and paired improvement on affected tasks. Appendix~\ref{app:paired_acceptance} details the classification and counts.

\vspace{-3pt}

\vspace{-2pt}
\subsection{Results on task-side execution efficiency}
\label{sec:ordinary_gui_results}
\vspace{-2pt}

\textbf{The shared runtime reduces task-side execution costs.}
Relative to MobileAgent, \frameworkname{} reduces mean E2E by 68.47\%--80.01\% and mean total-token usage by 41.86\%--59.56\% across the six AndroidWorld backbones. It also attains higher success on every backbone in this comparison. With GPT-5.5, E2E decreases from 838.03 to 196.88 seconds and success rises from 40.52\% to 72.4\%. With Gemini 3.1 Pro Preview, E2E decreases from 1008.0 to 272.96 seconds and success rises from 31.03\% to 61.2\%. These results support the applicability of the shared runtime to general application workflows. Appendix~\ref{app:aw_extended_analysis} gives the extended comparison.

% Main Table 2: application execution, with all three metrics.
\textbf{Fast termination alone does not imply efficient completion.} Relative to AppAgent, GPT-5.5 improves success by 37.1 percentage points based on the displayed SR values while reducing E2E by 30.23\%; GLM-5v-turbo also improves both success and E2E. With Claude Opus 4.8, higher success is accompanied by both higher E2E and greater total-token usage. With MiniMax-M3, E2E is lower, although total-token usage is higher. The native UI-TARS-1.5-7B system averages 95.24 seconds and 14,018.81 tokens, but completes 23 of 116 tasks. These comparisons show why short outputs and short runs must be considered together with task completion. \frameworkname{} improves completion rates and, in multiple system comparisons, reduces the time required for task execution.

\begin{table}[t]
    \vspace{-6pt}
    \centering
    \caption{Runtime-control ablation over 225 episodes per configuration. Each metric reports its value and, in gray, the change $\Delta$ relative to full supervision. SR differences are in percentage points; E2E and token differences use seconds and thousands of tokens, respectively.}
    \label{tab:runtime_ablation_main}
    \small
    \setlength{\tabcolsep}{5pt}
    \renewcommand{\arraystretch}{1.2}

    \resizebox{0.9\linewidth}{!}{%
    \begin{tabularx}{\linewidth}{@{}>{\raggedright\arraybackslash}X r@{\hspace{5pt}}l r@{\hspace{5pt}}l r@{\hspace{5pt}}l@{}}
        \toprule
        \textbf{Runtime configuration} & \multicolumn{2}{c}{\textbf{SR} (\%)} & \multicolumn{2}{c}{\textbf{E2E} (s)} & \multicolumn{2}{c}{\textbf{Tok} (k)} \\
        \midrule
        \rowcolor{oursbg}
        Full supervision & \textbf{82.22} & & \textbf{112.13} & & \textbf{41.82} & \\
        No complete mediation & 72.89 & \textcolor{gray!70}{\scriptsize $-$9.33} & 123.92 & \textcolor{gray!70}{\scriptsize $+$11.79} & 48.09 & \textcolor{gray!70}{\scriptsize $+$6.27} \\
        No fail-safe defaults & 75.11 & \textcolor{gray!70}{\scriptsize $-$7.11} & 158.13 & \textcolor{gray!70}{\scriptsize $+$46.00} & 60.64 & \textcolor{gray!70}{\scriptsize $+$18.82} \\
        No binding & 71.56 & \textcolor{gray!70}{\scriptsize $-$10.67} & 128.71 & \textcolor{gray!70}{\scriptsize $+$16.58} & 49.32 & \textcolor{gray!70}{\scriptsize $+$7.50} \\
        Relaxed supervisor & 68.44 & \textcolor{gray!70}{\scriptsize $-$13.78} & 119.97 & \textcolor{gray!70}{\scriptsize $+$7.84} & 46.84 & \textcolor{gray!70}{\scriptsize $+$5.02} \\
        Unbounded recovery & 83.56 & \textcolor{gray!70}{\scriptsize $+$1.33} & 417.45 & \textcolor{gray!70}{\scriptsize $+$305.32} & 101.40 & \textcolor{gray!70}{\scriptsize $+$59.58} \\
        \bottomrule
    \end{tabularx}%
    }

    \vspace{-12pt}
\end{table}
\FloatBarrier

\subsection{Ablating runtime controls}
\label{sec:runtime_ablation}

The ablation uses GPT-5.5 on 25 instances from each of nine HLL families, yielding 225 episodes per configuration. We compare full control against variants that bypass mediation, remove fail-safe defaults, relax binding with or without automatic submission, or remove recovery caps.

\textbf{Binding contributes under fixed automatic submission.}
Relaxing binding alone reduces success from 185/225 (82.22\%) to 161/225 (71.56\%), while mean E2E rises from 112.13 to 128.71 seconds and tokens from 41.82k to 49.32k. Because automatic submission is unchanged, this comparison isolates the contribution of binding in this configuration.

\textbf{Weaker runtime control reduces success while increasing cost.}
Bypassing mediation lowers success by 9.33 percentage points, while jointly relaxing binding and submission lowers it by 13.78 points. Removing fail-safe defaults reduces success by 7.11 points and produces the largest cost increase, adding 46.00 seconds and 18,818 tokens per run. These results support coordinating execution with available evidence rather than allowing unconstrained dispatch.

\textbf{Recovery caps provide a success--cost trade-off.}
Removing only the recovery caps increases success from 185/225 to 188/225, but raises E2E and token usage to 3.7$\times$ and 2.4$\times$ their full-control values. Thus, additional recovery yields only a small completion gain at substantially higher cost, supporting bounded recovery. Appendix~\ref{app:runtime_ablation_details} provides full configurations and analyses.

\vspace{-3pt}
\section{Conclusion}
\label{sec:conclusion}
\vspace{-3pt}

We studied GUI task execution in \emph{protected environments}, where goal achievement and environmental acceptance are distinct. \frameworkname{} combines a hierarchical solving tree with a runtime supervisor that treats tool outputs as revisable evidence, binds them to action commitments, mediates dispatch, and budgets recovery by category. Across evaluated backbones, it improves success in most verification settings and all AndroidWorld comparisons while reducing latency in several cases. A paired GPT-5.5 diagnostic further suggests gains when agents find correct solutions but fail to execute them acceptably, although improvements are not uniform. Our verification evaluation uses recorded interaction evidence rather than production detectors, while AndroidWorld measures general workflow execution. Extending these results to deployed protections and identifying the backbone properties underlying runtime supervision remain future work.

% ICLR 强制/推荐声明：必须排在正文结束后、参考文献之前（不占正文页数限制）。
\subsection*{AI use statement.}
Large language models served both as systems under evaluation and as writing assistants for manuscript organization, drafting, and language editing. The authors retain full responsibility for the accuracy and integrity of this work.

\subsection*{Ethics statement}
This work evaluates GUI agents in protected environments with CAPTCHA and behavioral anti-automation checks. All experiments are conducted in controlled settings using the locally deployed HLL verification environments and the AndroidWorld simulator; no real third-party websites, production verification systems, or user data are involved.
Our goal is to study the gap between solving a verification challenge and satisfying environmental checks, not to bypass deployed protections. We acknowledge that techniques improving interaction success could potentially be misused against real anti-automation systems. Our evaluation relies only on interaction evidence recorded by HLL rather than production detectors. The implementation will be released upon acceptance. The authors declare no competing interests.

\subsection*{Reproducibility statement}
Environment configuration, initialization, and execution budgets are documented in Appendices~\ref{app:env_init} and~\ref{app:budgets_termination}. Agent integration and information-access boundaries are described in Appendices~\ref{app:agent_integration} and~\ref{app:interface_boundaries}, while metric aggregation and ablation settings are given in Appendices~\ref{app:metric_aggregation} and~\ref{app:ablation_controls}. The full problem formulation and execution protocol are provided in Appendices~\ref{app:problem_formulation} and~\ref{app:execution_protocol}.
% 参考文献：条目保存在项目根目录的 iclr2027_conference.bib 中。
% 排版位置：正文结束后、附录之前。
\bibliography{iclr2027_conference}
\bibliographystyle{iclr2027_conference}

% 附录：内容写在 sections/appendix.tex 中，排在参考文献之后。
\appendix
\clearpage
\section{Extended Problem Formulation}
\label{app:problem_formulation}
\label{sec:appendix}

This appendix develops the protected-environment task model in Section~\ref{sec:protected_gui_tasks}: interaction dynamics, evidence-dependent verification, task-level acceptance, and end-to-end evaluation. It describes the environment in which an agent operates, not the internal rules of our runtime supervisor. Tool interfaces and supervisor decisions are specified separately in Appendix~\ref{app:runtime_interfaces}.

\subsection{Protected interaction dynamics and observations}
\label{app:protected_dynamics}

Given a user goal $g$, the agent receives an observation $o_t$ at step $t$ and executes a GUI action $a_t \in \mathcal{A}_{\mathrm{GUI}}$. Let $s_t \in \mathcal{S}$ denote the internal environment state, including the application state and verification-related context. Let $e_t$ denote the execution evidence actually produced by the action, including interaction events, timing information, and relevant state changes. The vector $\mathbf{v}_t = (v_t^{(1)}, \ldots, v_t^{(K)})$ records the protection outcomes at that step. A terminated task execution is recorded as
\begin{equation}
    \begin{aligned}
        \tau^{\mathrm{prot}} = \big(
        &s_0, o_0, a_0, e_0, \mathbf{v}_0,
        s_1, o_1, \ldots, \\
        &a_{T-1}, e_{T-1}, \mathbf{v}_{T-1},
        s_T, o_T
        \big),
    \end{aligned}
    \label{eq:protected_gui_trajectory}
\end{equation}
where $T$ is the number of interaction steps before termination. This representation distinguishes the action executed by the agent, $a_t$, the evidence actually produced by that action, $e_t$, and the environment's determinations based on that evidence, $\mathbf{v}_t$. A GUI action may generate multiple fine-grained interaction events, so actions and execution evidence need not have a one-to-one correspondence.

Let $H_t^{\mathrm{env}}$ denote the prefix of this record ending at $o_t$, before $a_t$ is included. A single interaction step in the protected environment is represented as
\begin{equation}
    \big(e_t, \mathbf{v}_t, s_{t+1}, o_{t+1}\big)
    \sim P^{\mathrm{prot}}
    \big(\,\cdot \mid H_t^{\mathrm{env}}, a_t\big),
    \label{eq:protected_gui_transition}
\end{equation}
where $P^{\mathrm{prot}}$ describes the interaction dynamics jointly induced by action execution, protection outcomes, and environmental feedback. Protection mechanisms therefore do not merely evaluate an execution after it ends: their determinations can also affect subsequent environment states and task progress.

The trajectory $\tau^{\mathrm{prot}}$ is an environment-side record used to define the task; it does not imply that all its variables are visible to the agent. At the GUI-interaction level, the available observation--action history is
\begin{equation}
    h_t = (o_0, a_0, o_1, \ldots, a_{t-1}, o_t).
    \label{eq:agent_visible_history}
\end{equation}
The goal $g$ and the information in this history inform subsequent interaction. Here, $o_t$ may include screenshots, accessibility information, and exposed feedback, including results from intervening tool calls. This GUI-level history does not prescribe how individual model or tool calls are scheduled. We do not assume that the agent can access internal environment states, complete execution evidence, or undisclosed protection outcomes.

The environment-side prefix $H_t^{\mathrm{env}}$ is distinct from the agent-visible history $h_t$. Neither an unavailable state nor an undisclosed verification outcome is added to $h_t$ merely because it appears in $\tau^{\mathrm{prot}}$.

\subsection{Protection evidence and verification rules}
\label{app:verification_rules}

For protection mechanism $j$, let $\mathcal{Z}_j$ denote its evidence space. The evidence used by this mechanism at step $t$ is
\begin{equation}
    z_t^{(j)} = \phi_j(H_t^{\mathrm{env}}, a_t, e_t)
    \in \mathcal{Z}_j.
    \label{eq:protection_evidence}
\end{equation}
The evidence-extraction function $\phi_j$ may select only the execution record of the current action, a local interaction window, or a longer execution history. This definition accommodates both single-step checks and cross-step verification without requiring every protection mechanism to depend on the complete trajectory.

Let $\mathcal{B}_j(s) \subseteq \mathcal{Z}_j$ denote the set of evidence satisfying verification requirement $j$ in environment state $s$. The corresponding verification rule is
\begin{equation}
    \begin{aligned}
        \mathcal{V}_j &: \mathcal{S} \times \mathcal{Z}_j
        \longrightarrow \{0,1\}, \\
        \mathcal{V}_j(s,z) &= \mathbf{1}\big[z \in \mathcal{B}_j(s)\big].
    \end{aligned}
    \label{eq:verification_rule}
\end{equation}
An output of $1$ indicates that the checked interaction passes the rule, whereas $0$ indicates that it does not. Let $\mathcal{J}_t \subseteq \{1,\ldots,K\}$ index the mechanisms that produce a determination at step $t$. Then
\begin{equation}
    v_t^{(j)} =
    \begin{cases}
        \mathcal{V}_j\big(s_t, z_t^{(j)}\big), & j \in \mathcal{J}_t, \\[2mm]
        \perp, & j \notin \mathcal{J}_t,
    \end{cases}
    \label{eq:verification_outcome}
\end{equation}
where $\perp$ means that no new determination is produced at that step, rather than that verification has passed. The joint transition law in Eq.~\ref{eq:protected_gui_transition} incorporates these outcomes together with their effects on the application and verification state.

This representation covers two distinct verification objectives. For interaction-validity checks, $\mathcal{B}_j(s)$ may be determined by the permissible range of an individual action, consistency across consecutive actions, or task-specific state-transition requirements. For human-versus-automation discrimination, let $D_j(z) \in [0,1]$ denote a discriminator's human-behavior score. One possible threshold-based acceptance rule is
\begin{equation}
    \mathcal{V}_j(s,z) = \mathbf{1}\big[D_j(z) \geq \theta_j\big],
    \label{eq:behavioral_verification}
\end{equation}
where $\theta_j \in [0,1]$ is a threshold set by the environment. The former class checks whether interactions satisfy task requirements; the latter assesses whether behavior exhibits distinguishable characteristics of automation. Both can affect task execution through environmental acceptance conditions. These rules characterize how the environment makes determinations; their internal parameters, scores, and acceptance sets are not assumed to be exposed to the agent.

\subsection{Environmental acceptance and task completion}
\label{app:trace_requirements}

The goal predicate $\mathrm{Goal}_g(\tau^{\mathrm{prot}})$ records whether the user-requested outcome is achieved. Environmental acceptance additionally depends on the verification requirements for the current task. Let $\mathcal{S}_{g,\mathcal{V}}^{\mathrm{acc}}\subseteq\mathcal{S}$ be the states in which these requirements have been satisfied:
\begin{equation}
    \mathrm{Accept}_{g,\mathcal{V}}\big(\tau^{\mathrm{prot}}\big)
    =\mathbf{1}\left[s_T\in\mathcal{S}_{g,\mathcal{V}}^{\mathrm{acc}}\right].
    \label{eq:environmental_acceptance}
\end{equation}
This terminal-state representation assumes that $s_T$ retains the verification context needed to determine acceptance for this execution. When acceptance depends on earlier interactions, the state includes the relevant verification history or its sufficient summary, rather than only the final visible page. The accepting set is task-dependent: it concerns the mechanisms required for completing $g$, not every mechanism deployed in the environment.

Local outcomes $v_t^{(j)}$ and overall acceptance have different roles. A local rejection can leave the task open to correction, while a local pass need not satisfy later verification requirements. If retries are permitted, the environment may record a rejection, a renewed attempt, and eventual entry into an accepting state. If a rejection irreversibly blocks the workflow, no such continuation is available. Likewise, $\perp$ is not a pass and does not by itself establish satisfaction of a required check. Overall acceptance is therefore not defined as the conjunction of all earlier local checks.

Equation~\ref{eq:protected_task_success} combines environmental acceptance with goal achievement; neither a local pass nor acceptance alone implies completion of the user's request. The two predicates need not be independent: passing a verification gate can itself be necessary to reach the goal. They define evaluation conditions, not an oracle exposed to the agent. The runtime supervisor's decision to dispatch an operation is separate from the environment's decision to accept the resulting interaction.

\subsection{End-to-end latency and evaluation}
\label{app:timing}

Each execution $\tau^{\mathrm{prot}}$ is associated with timing metadata and a termination reason. Its start timestamp records the beginning of the first model call for the task, and its end timestamp records success or failure termination. These timestamps include computation and waiting between GUI interactions, not just the action events in the trajectory. Consistent with Eq.~\ref{eq:end_to_end_latency},
\begin{equation}
    L(\tau^{\mathrm{prot}})
    =t_{\mathrm{end}}(\tau^{\mathrm{prot}})-t_{\mathrm{start}}(\tau^{\mathrm{prot}}).
    \label{eq:appendix_e2e}
\end{equation}
The interval covers model inference, tool calls, runtime control, GUI execution, environmental responses, verification, and recovery. It is measured as elapsed wall-clock time, not a sum of component durations that might overlap. For an execution stopped by an external cutoff, the interval ends at the actual cutoff; it does not estimate the unobserved time to eventual success. Failure and cutoff runs retain their elapsed time and termination status.

For a fixed task distribution and evaluation interface, we distinguish
\begin{equation}
    \begin{aligned}
        P_{\mathrm{succ}}&=\mathbb{E}[S_g(\tau^{\mathrm{prot}})],\\
        \overline{L}_{\mathrm{all}}&=\mathbb{E}[L(\tau^{\mathrm{prot}})],\\
        \overline{L}_{\mathrm{succ}}&=\mathbb{E}[L(\tau^{\mathrm{prot}})\mid S_g(\tau^{\mathrm{prot}})=1].
    \end{aligned}
    \label{eq:appendix_metrics}
\end{equation}
The expectation is over task instances and any randomness in the agent and environment. All-run latency includes failed executions; successful-run latency describes only completed tasks and is undefined when there are no successes. Because methods may succeed on different task subsets, their conditional means do not alone establish a speedup on the same workload. Timing comparisons must identify the population being compared and report success rates and termination conditions alongside latency.

These measures distinguish reliable acceleration from early failure. A lower all-run latency accompanied by more failed tasks is not evidence of faster successful completion. Token usage and action counts may help explain execution cost, but do not replace wall-clock measurement. Evaluation cutoffs are external limits; the framework's recovery budgets and their accounting are described in Appendix~\ref{app:recovery_details}.
\section{Execution Protocol Details}
\label{app:execution_protocol}

This appendix expands the operational semantics of the solving tree and runtime supervisor in Section~\ref{sec:method}. The predicates specify conditions for capability entry, evidence binding, and progress assessment; task interfaces provide the concrete checks and completion signals. They do not imply complete visual verification or optimal control. The task examples illustrate protocol semantics rather than report measured trajectories. Notation follows the main text: $k$ indexes runtime decisions, and $\mathcal{H}_k$ records accessible attempt outcomes.

\subsection{Capability instances and dependencies}
\label{app:capability_details}

The hierarchical solving tree is built from a reusable capability skeleton. A task adapter supplies the goal, available affordances, supported tools, and evidence or completion interfaces, thereby determining which parts of the skeleton apply. Task conditioning does not require generating new capability types or searching over parallel candidate plans.

The node specification in Eq.~\ref{eq:solving_tree} uses $\mathrm{Pre}_n(\Sigma)=\mathrm{Pre}(n,\Sigma)$, $\mathrm{Act}_n=\mathrm{Act}(n)$, and $\mathrm{Done}_n(\Sigma)=\mathrm{Done}(n,\Sigma)$. Here, Pre and Done are Boolean predicates, and Act is a set of operations. Let $\mathrm{Pred}_g(n)$ denote prerequisite nodes in the task's dependency specification. The semantic readiness condition is
\begin{equation}
    \mathrm{Ready}(n,\Sigma)
    =\mathrm{Pre}(n,\Sigma)\land
      \bigwedge_{m\in\mathrm{Pred}_g(n)}\mathrm{Done}(m,\Sigma).
    \label{eq:capability_readiness}
\end{equation}
For an AND node with required children $\mathrm{Child}_{\land}(n)$, completion requires
\begin{equation}
    \mathrm{Done}(n,\Sigma)
    =\bigwedge_{m\in\mathrm{Child}_{\land}(n)}\mathrm{Done}(m,\Sigma).
    \label{eq:and_completion}
\end{equation}
An OR branch offers alternative ways to satisfy a capability; selecting a tool is not itself evidence of capability completion. Ready describes the intended execution dependencies. Its implementation can combine runtime checks with model guidance without constructing a complete Boolean mask for all nodes.

Completion evidence is interpreted in the current task context. A changed target or refreshed challenge can invalidate earlier grounding while leaving unrelated evidence intact. Revisiting Ground or Solve is an execution transition over the capability structure rather than a new search-tree expansion. Verify can be entered after a failed or unresolved attempt: progress evidence is a condition for advancement, not a prerequisite for inspecting an outcome.

\begin{table}[t]
    \centering
    \caption{Capability roles and their advancement requirements. Roles denote semantic responsibilities rather than necessarily separate model calls, tools, or modules.}
    \label{tab:capability_obligations}
    \small
    \setlength{\tabcolsep}{6pt}
    \renewcommand{\arraystretch}{1.18}
    \begin{tabularx}{\linewidth}{@{}>{\raggedright\arraybackslash}p{0.12\linewidth}>{\raggedright\arraybackslash}X>{\raggedright\arraybackslash}X@{}}
        \toprule
        \textbf{Capability} & \textbf{Context \& operation} & \textbf{Evidence for advancement} \\
        \midrule
        \cellcolor{stagebg}\textbf{Perceive} & Task goal and observation access; inspect the current page. & Relevant objects and affordances are identified sufficiently for subsequent reasoning. \\
        \addlinespace[2pt]
        \cellcolor{stagebg}\textbf{Ground} & Resolve a target or region; localize, associate, or match objects. & Target geometry, identity, correspondence, or ordering is established. \\
        \addlinespace[2pt]
        \cellcolor{stagebg}\textbf{Solve} & Reason over grounded task content to infer an answer or candidate action. & A supported solution hypothesis or actionable result is available. \\
        \addlinespace[2pt]
        \cellcolor{stagebg}\textbf{Act} & Execute an applicable operation under the current stage constraints. & The environment returns an execution outcome requiring assessment. \\
        \addlinespace[2pt]
        \cellcolor{stagebg}\textbf{Verify} & Inspect feedback after an attempted action or unresolved outcome. & Evidence supports progress, completion, failure, or continued uncertainty. \\
        \bottomrule
    \end{tabularx}
\end{table}

Ground and Solve can be fulfilled by one tool call when its output supplies both target information and a candidate operation. A simple GUI task may not need a separate solving invocation. The abstraction preserves evidence dependencies without requiring each semantic role to correspond to a distinct software module.

\subsection{Evidence support and binding conditions}
\label{app:evidence_details}

The retained runtime evidence $E_k$ consists of accessible tool results and observations supporting execution decisions; it is distinct from the environment-side execution evidence $e_t$ in Appendix~\ref{app:protected_dynamics}, which need not be fully exposed. An accessible evidence item can be described semantically as
\begin{equation}
    \eta=(\mathrm{source},\mathrm{target},\mathrm{content},
          \mathrm{scope},\mathrm{context}),
    \label{eq:evidence_description}
\end{equation}
where context associates the result with the observation or task context in which it was obtained. Evidence sources need not expose identical fields, explicit state identifiers, or numerical freshness scores. Applicability depends on whether accessible information supports the relevant assumptions, rather than simply on how recently the evidence was produced.

For a candidate operation $a^c$ supported by evidence $E^c\subseteq E_k$, the runtime establishes a commitment when the binding checks pass:
\begin{equation}
    c_k\leftarrow(n,a^c,E^c)
    \quad\text{when}\quad
    \mathrm{BindOK}(n,a^c,E^c,\Sigma_k)=1.
    \label{eq:evidence_binding}
\end{equation}
Here, $n$ is the capability served by the operation. The commitment authorizes a continuation under the supported checks; it does not certify the tool's semantic correctness.

Let StageOK denote compatibility with the active stage, Covers denote evidence support for the operation's target, and $\mathrm{Check}_{\mathrm{ctx}}$ denote the contextual checks supported by the task interface. BindOK can be abstractly factored as
\begin{equation}
    \begin{aligned}
        &\mathrm{BindOK}(n,a,E^c,\Sigma)\\
        &\quad=\mathrm{StageOK}(n,a,\Sigma)
        \land\mathrm{Covers}(E^c,a)\\
        &\qquad\land
        [\mathrm{Check}_{\mathrm{ctx}}(E^c,\Sigma)=\mathrm{pass}].
    \end{aligned}
    \label{eq:binding_checks}
\end{equation}
The result pass refers only to the checks the interface supports, not an exhaustive test of evidence truth. A failed or unresolved required check does not justify binding; missing information may instead prompt renewed observation or grounding. Dispatch remains subject to the applicable stage and budget constraints even after a binding has been established.

Evidence can support several decisions, but a pending commitment records one specific executable continuation. This relationship distinguishes supplying a tool result as model context from having the runtime enforce the operation that the result supports.

\paragraph{Reconsidering evidence.} A new challenge, a changed target region, an unavailable object, or feedback contradicting the current solution can trigger reassessment of dependent commitments. Content rejection can undermine a solution hypothesis without invalidating every geometric observation; a changed page can invalidate geometry while leaving the high-level goal intact. Budget exhaustion restricts execution but does not itself make evidence false. The checks supported by the interface determine which changes can be detected; the protocol does not assume an oracle for all visual changes.

\subsection{Commitment lifecycle and decision resolution}
\label{app:commitment_lifecycle}

The commitment lifecycle has four logically distinct situations:
\begin{enumerate}
    \item \textbf{Unbound.} Evidence is insufficient to establish the next committed operation. The model can propose relevant evidence acquisition or another permitted operation.
    \item \textbf{Pending.} Actionable evidence establishes $c_k$. In the full-control configuration, a still-admissible commitment takes precedence over a conflicting proposal.
    \item \textbf{Dispatched, outcome unresolved.} The operation has been issued and is no longer pending for repetition. Its effect still requires assessment through feedback or observation.
    \item \textbf{Resolved or reconsidered.} Progress supports advancement; failure or changed support motivates revising the hypothesis, reacquiring evidence, or recovering within budget.
\end{enumerate}
These are semantic distinctions rather than four required implementation flags. Consuming the pending status and recording an attempt must preserve the information needed for outcome verification.

For the controller in Eq.~\ref{eq:framework_decision}, replacing a proposal is justified by a currently admissible commitment, not merely by an old pending record. Admissibility assessment must incorporate the latest accessible feedback. A commitment cannot override terminal conditions or violated enforced constraints. When its support no longer holds, it loses its basis for execution precedence, and the runtime reconsiders the continuation.

Rejecting a proposal, deferring an interaction, and initiating recovery need not each produce a new GUI action. A rejected proposal can be withheld; a deferred GUI interaction can be replaced by a grounding or verification request; recovery can revise the active capability before the next interaction is selected. A $\mathrm{hold}$ decision dispatches no request, so active evidence acquisition must be selected as an operation in $\mathcal{A}$. All dispatched replacements remain subject to applicable checks. These dependencies do not require every implementation to use the same ordered list of conditional branches.

\begin{table}[t]
    \centering
    \caption{State-conditioned treatment of re-localization proposals. The cases illustrate protocol behavior rather than an exhaustive rule-priority list.}
    \label{tab:protocol_decisions}
    \small
    \setlength{\tabcolsep}{6pt}
    \renewcommand{\arraystretch}{1.18}
    \begin{tabularx}{\linewidth}{@{}>{\raggedright\arraybackslash}p{0.36\linewidth}>{\raggedright\arraybackslash}X@{}}
        \toprule
        \textbf{Current protocol state} & \textbf{Runtime treatment} \\
        \midrule
        \cellcolor{statebg}\textbf{Target evidence is missing} & Allow acquisition of relevant evidence. \\
        \addlinespace[2pt]
        \cellcolor{statebg}\textbf{An admissible commitment is pending} & Prioritize the committed operation under full control. \\
        \addlinespace[2pt]
        \cellcolor{statebg}\textbf{Supporting conditions have changed} & Reconsider the current binding and acquire relevant evidence. \\
        \addlinespace[2pt]
        \cellcolor{statebg}\textbf{Repeated tool failure without new evidence} & Redirect selection toward an alternative capability. \\
        \addlinespace[2pt]
        \cellcolor{statebg}\textbf{Applicable recovery budget is exhausted} & Stop recovery operations governed by that budget. \\
        \bottomrule
    \end{tabularx}
\end{table}

The same proposal can lead to different decisions in different states. An admissible pending commitment can replace repeated localization, whereas changed page conditions can make that localization necessary again. Proposal identity alone is therefore insufficient to determine the runtime response.

\subsection{Progress assessment and feedback-directed recovery}
\label{app:recovery_details}

For nonterminal decisions, the runtime assesses outcomes using the active stage, protocol state, actual decision, and subsequent accessible feedback:
\begin{equation}
    \begin{aligned}
        \rho_k
        &=\mathrm{Assess}
        (q_k,\Sigma_k,d_k,f_k,\bar{o}_{k+1}),\\
        \rho_k
        &\in\{\mathrm{progress},\mathrm{failure},\mathrm{unresolved}\}.
    \end{aligned}
    \label{eq:progress_assessment}
\end{equation}
Here, $f_k$ is the feedback accessible after the decision, and $\bar{o}_{k+1}$ is the latest available observation. The categories represent evidence of progress, evidence against the current attempt, and a lack of decisive evidence. They need not be literal labels returned by the implementation or environment. An unresolved outcome is not a success, and a visible change need not constitute useful progress; completion signals must match the active capability or task criterion. Runtime outcome verification assesses accessible evidence; it does not replace the environment's protection checks or guarantee environmental acceptance.

Equation~\ref{eq:protocol_update} couples updates to the stage, evidence, commitment, outcome history, and budgets. An operation can consume a commitment without completing the goal; rejection can revise a hypothesis without abandoning the task; and a tool call can produce evidence without changing the GUI. A single generic success flag cannot represent these distinctions.

\begin{table}[t]
    \centering
    \caption{Default recovery directions conditioned on observable feedback. They guide local recovery decisions rather than define a complete failure diagnosis or cost-optimal policy.}
    \label{tab:recovery_directions}
    \small
    \setlength{\tabcolsep}{6pt}
    \renewcommand{\arraystretch}{1.18}
    \begin{tabularx}{\linewidth}{@{}>{\raggedright\arraybackslash}p{0.27\linewidth}>{\raggedright\arraybackslash}X@{}}
        \toprule
        \textbf{Feedback situation} & \textbf{Default recovery direction} \\
        \midrule
        \cellcolor{statebg}\textbf{Tool not applicable} & Reconsider the current affordance and select a relevant grounding or capability branch. \\
        \addlinespace[2pt]
        \cellcolor{statebg}\textbf{Tool error} & Seek another evidence source or reacquire the inputs required by the tool. \\
        \addlinespace[2pt]
        \cellcolor{statebg}\textbf{Behavioral rejection} & Reconsider the interaction form or sequence while retaining the solution hypothesis when justified. \\
        \addlinespace[2pt]
        \cellcolor{statebg}\textbf{Content rejection} & Revise the solution; correct locally when possible, otherwise refresh and ground again. \\
        \addlinespace[2pt]
        \cellcolor{statebg}\textbf{Verification unresolved} & Obtain another observation or progress signal before declaring success or repeating a submission. \\
        \addlinespace[2pt]
        \cellcolor{statebg}\textbf{Repeated failure without new evidence} & Use failed-attempt history to steer selection away from uninformative repetition. \\
        \addlinespace[2pt]
        \cellcolor{statebg}\textbf{Recovery budget exhausted} & Stop recovery operations governed by the budget; terminate when no permitted continuation remains. \\
        \bottomrule
    \end{tabularx}
\end{table}

Explicit environment responses can supply rejection categories. Other conditions, such as missing progress or repeated evidence, can be inferred from observations and history. The protocol does not assume access to hidden ground-truth failure causes. Branch selection combines runtime constraints with model reasoning guided by feedback and history. Shared state therefore constrains and guides recovery rather than hard-masking every unsuitable branch.

For budget component $b_k^{(j)}$, count-based accounting can be written as
\begin{equation}
    b_{k+1}^{(j)}=b_k^{(j)}-\Delta_k^{(j)},
    \qquad 0\leq\Delta_k^{(j)}\leq b_k^{(j)},
    \label{eq:budget_accounting}
\end{equation}
where $\Delta_k^{(j)}$ counts the recovery units charged to that component. Units are configuration-dependent, such as an attempt or refresh. This equation specifies accounting rather than numerical budget settings or measured overhead, and does not by itself bound all model calls or external waiting time. Re-entering a capability cannot automatically restore an exhausted episode-level budget. Any separate local-budget reset must be distinguished from the outer episode limit.

\subsection{Runtime interfaces and execution cycle}
\label{app:protocol_cycle}
\label{app:runtime_interfaces}

Let $\bar{o}_k$ denote the latest observation accessible to the runtime, $\pi_\theta$ the proposal policy, and $\psi$ the supervisor's rules and configuration. Tool calls and direct GUI actions form the dispatchable request set $\mathcal{A}=\mathcal{U}\cup\mathcal{A}_{\mathrm{GUI}}$. The model selects a capability and refines a request using its permitted state view; the supervisor decides what to dispatch, and feedback updates the protocol state:
\begin{align}
    n_k
    &=\mathrm{Select}_{\theta}
      (\mathcal{T}_g,\bar{o}_k,\Sigma_k^\pi),\notag\\
    \tilde{a}_k
    &=\mathrm{Refine}_{\theta}
      (n_k,\bar{o}_k,\Sigma_k^\pi),
    \label{eq:capability_proposal}\\
    d_k
    &=\mathcal{C}_{\psi}
      (g,\bar{o}_k,\Sigma_k,\tilde{a}_k),
    \label{eq:framework_decision}\\
    \Sigma_{k+1}
    &=\mathrm{Update}_{\psi}
      (\Sigma_k,\bar{o}_k,\tilde{a}_k,d_k,f_k,\bar{o}_{k+1}).
    \label{eq:protocol_update}
\end{align}
Select and Refine are logical steps and need not be separate model calls. When $d_k\in\mathcal{A}$, the runtime dispatches $a_k=d_k$; $\mathrm{hold}$ dispatches no request, and $\mathrm{stop}$ terminates the run. Update applies to nonterminal decisions. If a hold decision produces no new feedback or observation, $f_k$ may be empty and $\bar{o}_{k+1}$ may retain the latest observation; holding does not automatically supply new information.

Only GUI actions that are actually dispatched enter the environment interaction record defined in Section~\ref{sec:protected_gui_tasks}. A tool may return evidence alone or also execute GUI interactions, so tool calls and environment interactions need not correspond one-to-one. Thus, $k$ indexes runtime decisions rather than a fixed number of low-level GUI events.

These interfaces yield the following logical cycle, whose steps need not correspond one-to-one to software functions or model calls:
\begin{enumerate}
    \item \textbf{Read the execution context.} Use the state incorporating accessible observations and prior feedback to inspect current evidence, stage status, commitments, outcome history, and budgets.
    \item \textbf{Form a candidate continuation.} Select and refine a proposal using the task capability structure and the permitted model view. A tool-supported pending commitment remains available to the runtime even when the model proposes another operation.
    \item \textbf{Resolve execution.} Check terminal conditions and applicable constraints, reassess commitments, and accept, replace, or withhold the proposal. Missing evidence can lead to evidence acquisition rather than an interaction that directly changes task state.
    \item \textbf{Dispatch and obtain the outcome.} Execute only the selected request, retain its supporting context, and obtain feedback while distinguishing execution status from task progress.
    \item \textbf{Update state and advance or recover.} Use the outcome to update the protocol state, advance a capability, revise a commitment, acquire evidence, or choose a bounded recovery direction, then continue or terminate.
\end{enumerate}
The cycle connects model proposals, runtime dispatch, and subsequent outcome evidence rather than introducing an additional search algorithm. Logical binding does not imply that every continuation saves a model call; call scheduling and binding are distinct concerns.

\subsection{Task instantiation and illustrative trajectories}
\label{app:task_instantiation}

\paragraph{Slider interaction.} Suppose a grounding or solving tool returns a candidate displacement whose evidence supports a drag commitment. After dispatch, exposed feedback indicating content rejection leaves the high-level goal unfinished and provides evidence against the attempted displacement. Corrective grounding or solving can produce a revised displacement and a new commitment, and the subsequent interaction is verified. This example illustrates selective evidence retention and commitment revision, not a benchmark result or a controlled comparison with another agent.

\paragraph{Ordinary GUI interaction.} For a settings-change task, the adapter supplies the desired setting and an observable completion criterion. Perception identifies the relevant page, grounding identifies the control, evidence supports an operation on that control, and verification checks whether the desired state is observed. A page change before dispatch or an unclear effect prompts renewed evidence acquisition or further observation. The reusable structure is the dependency between evidence, commitment, execution, and verification; the target and concrete checks remain task-specific. This example does not report a completed task from a particular benchmark.

\paragraph{Negative cases and limitations.} Repeated interactions may still fail to produce useful progress when feedback is ambiguous or model-guided recovery selects poorly. A submission may be deferred because prerequisites are unavailable, yet the run can still end at a cutoff. Enforcing execution constraints does not provide a solver that always reaches the goal. Incorrect tool evidence and excessive control overhead can also offset the intended benefits.

Measured case studies should retain the initial context, model proposal, supporting evidence, runtime decision and reason, actual operation, feedback, state change, and final outcome. Operation-name normalization alone does not demonstrate semantic correction. Likewise, two trajectories do not establish a causal comparison without aligned task instances, information access, and execution conditions. Illustrative protocol examples and measured cases in the experimental appendix are therefore presented separately.
\section{Rational Metareasoning and Design Rationale}
\label{app:metareasoning_rationale}

This appendix expands the design principles in Section~\ref{sec:design_rationale}. Rational metareasoning provides a perspective on allocating processing effort; the concrete execution mechanisms are specified in Section~\ref{sec:method} and Appendix~\ref{app:execution_protocol}.

\subsection{Decision improvement and processing cost}

Rational metareasoning distinguishes the choice of external actions from the choice of computations that inform those actions. A computation is valuable insofar as its expected contribution to subsequent decision quality justifies its cost \citep{russell1991principles}. This perspective addresses both which computations to perform and when to stop deliberating and act. For example, \citet{hay2012selecting} formulate computation selection as a metalevel decision problem in which computational outcomes inform the eventual choice of an external action. Performing more computation is therefore not an objective in itself.

We borrow this benefit--cost perspective for model reasoning, tool-based solving, evidence acquisition, and outcome verification. These activities contribute to end-to-end latency, and their usefulness depends on how they support subsequent execution. They are not all purely internal computations: obtaining an observation may require an environment interaction, and a tool may both return information and execute GUI actions. We therefore retain the distinction between tool invocations and realized interactions, rather than identify every tool call with a metalevel computational action. The corresponding runtime interfaces and execution rules are described in Appendix~\ref{app:execution_protocol}.

\subsection{Implications for protected execution}

Reliable and efficient execution depends not only on answer correctness, but also on when to acquire information, act on an available solution, and revise an attempt after feedback. Additional reasoning and checks can improve execution decisions, but also incur latency. Acting prematurely can lead to rejection or rework, whereas repeatedly revisiting an already-supported solution can add cost without improving the next decision. The design objective is thus to coordinate these activities, rather than maximize reasoning effort or minimize interaction count in isolation.

Environmental acceptance further makes reassessment consequential. A correct answer can be followed by a rejected interaction, and dispatch can leave the intended effect unresolved. Conversely, when recovery is permitted, feedback from a failed attempt can support a useful continuation. These conditions motivate the following design choices.

\paragraph{Reusing an applicable result.} An evidence-grounded commitment preserves a specific supported continuation across successive proposals. It allows the runtime to act on an available result while the supporting conditions and applicable execution constraints remain satisfied, rather than reopen selection without a decision-relevant reason. Relevant changes in the target, page context, or feedback can instead justify renewed grounding or solving. This reuse rule neither certifies semantic correctness nor implies that further computation has no value merely because a candidate operation exists.

\paragraph{Checking consequential uncertainty.} An incorrect progress judgment can cause subsequent operations to rely on an unfinished prerequisite. Verification incurs a cost, but can prevent downstream rework by informing whether to advance, reassess, or continue observing. This motivates checks tied to capability and task criteria, not a separate verification call after every internal computation. The objective is to obtain evidence relevant to continuation while satisfying the applicable verification requirements, rather than add checks indiscriminately or omit required verification to reduce measured time.

\paragraph{Using feedback within a budget.} Recovery revises assumptions undermined by feedback while retaining evidence that remains applicable. A rejected solution need not invalidate every geometric observation; a changed page may require renewed grounding without changing the user goal. Feedback and execution history should therefore guide subsequent attempts toward unresolved issues rather than mechanical repetition. Budgets limit designated recovery operations, as specified in Appendix~\ref{app:recovery_details}. These limits constrain repeated attempts but do not, by themselves, establish a bound on all model calls or external waiting time.

\subsection{Scope of the theoretical connection}

The three principles are design choices inspired by rational metareasoning, not theorems derived from an optimal metalevel policy. The framework implements them through accessible evidence, execution history, and budget constraints, without requiring an online estimator of expected computation value or a complete probabilistic model of computational outcomes. Neither the hierarchical solving tree nor the runtime supervisor is claimed to maximize a computation-value objective. Their effects on success and end-to-end latency are evaluated empirically under the timing and acceptance definitions in Appendix~\ref{app:problem_formulation}.
\section{Supplementary Experimental Details and Analysis}
\label{app:additional_experimental_results}

This appendix provides experimental configurations, extended quantitative analysis, statistical stability analysis, and a detailed examination of runtime ablations. A paired analysis separates static correctness from dynamic acceptance. Failure mode analysis covers twelve baseline cases and two text-based \frameworkname{} case studies connecting runtime decisions to successful and failed outcomes.

\subsection{Experimental details}
\label{app:experiment_details}

\subsubsection{Evaluation environments and episode initialization}
\label{app:env_init}

\paragraph{Dynamic verification.} The CAPTCHA evaluation runs in Google Chrome (\texttt{com.android.chrome}) on an Android Virtual Device identified as \texttt{agent-debug} (device serial \texttt{emulator-5554}). The emulator uses an ARM64-v8a Google Play system image at Android API level~36 and is configured in portrait orientation at $1080\times2400$ pixels and 420~dpi, with 4~vCPUs and 2048~MB of RAM. ADB opens each locally hosted CAPTCHA page and performs screenshots, text entry, taps, and drag gestures. Each of the nine HLL task families provides 100 instances; the backbone--harness comparisons in Table~\ref{tab:hll_main} score 90 instances in total, while the native UI-TARS-1.5-7B evaluation in Appendix~\ref{app:ui_tars_native_results} scores the full 900-instance population. Episodes use distraction level~2 and \texttt{hard} difficulty where the task family supports it, otherwise \texttt{normal}; the distraction level does not specify a fixed number of distractors. Each episode creates an independent local session from its task instance and seed, resets the evaluation state, and destroys the associated state after termination.

\paragraph{Application workflows.} AndroidWorld evaluates application-level task completion on a Pixel~6 emulator running Android~13/API~33 with an ARM64 image at $1080\times2400$ pixels. Tasks use the official initialization, terminal evaluation, and clean-up lifecycle. The runtime ablation is a separate batch, containing 25 instances of each CAPTCHA type and 225 episodes per configuration.

\subsubsection{Agent integration}
\label{app:agent_integration}

\paragraph{Preserved execution procedures.} Each evaluated approach retains its own model prompting, interaction history, action interpretation, and device-level execution procedure. Framework-driven approaches retain their native planning, memory, reflection, and feedback mechanisms when these mechanisms are part of the evaluated method. UI-TARS Harness denotes the UI-TARS-style observation--action procedure paired with the same general-purpose backbone used by the other harnesses, whereas UI-TARS-1.5-7B denotes the native GUI policy with its associated harness. These two configurations retain separate records and result entries.

\paragraph{Common evaluation boundary.} The common evaluation layer initializes the task, provides the public task objective, applies the evaluation-wide limits, records the trajectory, invokes the terminal evaluator, and performs clean-up. It does not inspect or rewrite intermediate baseline actions, add method-specific recovery behavior, or expose private evaluation state. This separation preserves each method's execution procedure while holding the task lifecycle and final success determination fixed.

\subsubsection{Observation, action, and information boundaries}
\label{app:interface_boundaries}

\begin{table}[!htbp]
    \centering
    \caption{Agent-visible inputs, permitted interactions, and returned feedback in the CAPTCHA evaluation.}
    \label{tab:captcha_agent_interfaces}
    \footnotesize
    \setlength{\tabcolsep}{4.5pt}
    \renewcommand{\arraystretch}{1.16}
    \begin{tabularx}{\linewidth}{@{}>{\raggedright\arraybackslash}p{0.16\linewidth}*{3}{>{\raggedright\arraybackslash}X}@{}}
        \toprule
        \textbf{Method} & \textbf{Decision input} & \textbf{Permitted interaction} & \textbf{Returned feedback} \\
        \midrule
        \cellcolor{oursbg}\textbf{\frameworkname{}} & Current screenshot; public task instruction; retained interaction evidence; execution progress; enabled-tool outputs. & GUI actions plus enabled perception, localization, cropping, checking, refresh, and recovery capabilities. & Updated screenshot, public page changes, and feedback that does not reveal the answer. \\
        \addlinespace[2pt]
        \cellcolor{labelbg}\textbf{MobileAgent / AppAgent} & Native public observations and retained interaction history. & Native GUI action space and enabled capabilities. & Updated interface state and public environmental feedback. \\
        \addlinespace[2pt]
        \cellcolor{labelbg}\textbf{UI-TARS systems} & Task instruction, current screenshot, and limited interaction history. & Click, long press, drag, type, scroll, back, home, wait, screenshot, and termination actions. & Updated screenshot and terminal task outcome. \\
        \bottomrule
    \end{tabularx}
\end{table}
\FloatBarrier

\begin{table}[!htbp]
    \centering
    \caption{Evaluator-only information and success determination in the two evaluation environments.}
    \label{tab:evaluator_information_boundaries}
    \small
    \setlength{\tabcolsep}{6pt}
    \renewcommand{\arraystretch}{1.16}
    \begin{tabularx}{\linewidth}{@{}>{\raggedright\arraybackslash}p{0.16\linewidth}>{\raggedright\arraybackslash}X>{\raggedright\arraybackslash}p{0.27\linewidth}@{}}
        \toprule
        \textbf{Environment} & \textbf{Evaluator-only information} & \textbf{Success determination} \\
        \midrule
        \cellcolor{labelbg}\textbf{CAPTCHA} & Correct answer, hidden verification state, evaluator signals, and complete environment telemetry. & Independent environment verification of the executed interaction. \\
        \addlinespace[3pt]
        \cellcolor{labelbg}\textbf{AndroidWorld} & Private task parameters, the success oracle, and states or trajectories of other agents. & Official terminal evaluation of the resulting application state. \\
        \bottomrule
    \end{tabularx}
\end{table}
\FloatBarrier

\paragraph{Method-specific access.} UI-TARS systems receive neither OCR output, DOM or accessibility structures, \frameworkname{} tool results or execution state, nor verification answers. \frameworkname{} tools operate only on permitted interface observations and cannot query evaluator-only state. For every method, an agent-issued termination ends execution but does not itself establish success; the environment evaluates success independently.

\subsubsection{Information-isolation checks}
\label{app:information_isolation_checks}

\paragraph{Checks and scope.} Existing checks include tests of UI-TARS observation and dependency isolation and a separation-of-duties audit of model-visible observations, post-action feedback, and tool results. The inspected interfaces did not expose CAPTCHA answers or evaluator-only fields to the model. Deterministic perception and OCR components do not invoke a language model and therefore do not contribute language-model tokens. These checks cover defined interfaces and recorded fields; they are not a formal information-flow proof or evidence that every possible execution path has been exhaustively verified.

\subsubsection{Budgets, termination, and error handling}
\label{app:budgets_termination}

\begin{table}[!htbp]
    \centering
    \caption{Execution limits used in the evaluation.}
    \label{tab:execution_limits}
    \small
    \setlength{\tabcolsep}{6pt}
    \renewcommand{\arraystretch}{1.14}
    \begin{tabularx}{\linewidth}{@{}>{\raggedright\arraybackslash}p{0.37\linewidth}>{\centering\arraybackslash}p{0.18\linewidth}>{\raggedright\arraybackslash}X@{}}
        \toprule
        \textbf{Limit} & \textbf{Value} & \textbf{Scope} \\
        \midrule
        \cellcolor{labelbg}\textbf{Maximum output per model call} & \textbf{4096 tokens} & All CAPTCHA evaluations. \\
        \addlinespace[1.5pt]
        \cellcolor{labelbg}\textbf{Model-request timeout} & \textbf{180 s} & Each model request in the CAPTCHA evaluation. \\
        \addlinespace[1.5pt]
        \cellcolor{labelbg}\textbf{Task-level timeout} & \textbf{1200 s} & All CAPTCHA evaluations. \\
        \addlinespace[1.5pt]
        \cellcolor{labelbg}\textbf{GUI-action budget} & \textbf{32} & \frameworkname{}/HLL runtime. \\
        \addlinespace[1.5pt]
        \cellcolor{labelbg}\textbf{Read-only-step budget} & \textbf{4} & \frameworkname{}/HLL runtime; counted separately. \\
        \addlinespace[1.5pt]
        \cellcolor{labelbg}\textbf{Grounding-stage step cap} & \textbf{3} & \frameworkname{}/HLL runtime. \\
        \addlinespace[1.5pt]
        \cellcolor{labelbg}\textbf{Execution-phase step cap} & \textbf{24} & \frameworkname{}/HLL runtime. \\
        \addlinespace[1.5pt]
        \cellcolor{labelbg}\textbf{Correction / refresh caps} & \textbf{1 / 1} & \frameworkname{}/HLL runtime, except the recovery-cap ablation. \\
        \addlinespace[1.5pt]
        \cellcolor{labelbg}\textbf{Maximum invalid actions} & \textbf{2} & Confirmed \frameworkname{}/HLL batches. \\
        \addlinespace[1.5pt]
        \cellcolor{labelbg}\textbf{AndroidWorld action budget} & \textbf{$10\times$ complexity} & AndroidWorld evaluations. \\
        \bottomrule
    \end{tabularx}
\end{table}
\FloatBarrier

\paragraph{Termination and reruns.} CAPTCHA success requires an actual successful verification by the environment, and AndroidWorld success requires the official terminal evaluator to accept the final state. Invalid model output, tool errors, task timeout, budget exhaustion, unrecoverable errors, and agent-requested termination without environmental success are recorded as failures. A zero-action infrastructure interruption may be retried only under a predetermined policy, with the original record retained; once a model action or environment interaction has occurred, a failed episode is not replaced by a later successful rerun.

\subsubsection{Episode selection and metric aggregation}
\label{app:metric_aggregation}

\paragraph{Scored populations.} Let $\mathcal{I}_{m,b}$ denote the final scored episode set for method $m$ with backbone $b$, and let $S_i$, $L_i$, and $O_i$ denote environmental success, observed end-to-end latency, and cumulative total tokens for episode $i$, respectively. We compute
\begin{equation}
    \mathrm{SR}_{m,b}=\frac{\sum_{i\in\mathcal{I}_{m,b}}S_i}{|\mathcal{I}_{m,b}|},\qquad
    \overline{\mathrm{E2E}}_{m,b}=\frac{\sum_{i\in\mathcal{I}_{m,b}}L_i}{|\mathcal{I}_{m,b}|},\qquad
    \overline{\mathrm{Tok}}_{m,b}=\frac{\sum_{i\in\mathcal{I}_{m,b}}O_i}{|\mathcal{I}_{m,b}|}.
    \label{eq:experiment_metric_aggregation}
\end{equation}
Each configuration uses the same scored population for all three metrics, including successful episodes, model failures, output-format failures, tool errors, and task timeouts. Successful-episode-only statistics, when reported, are labeled separately.

\paragraph{Timing and token accounting.} E2E is the elapsed wall-clock interval from the first model call to success or failure termination, as formalized in Appendix~\ref{app:timing}; it includes model inference, tool calls, runtime control, GUI execution, environmental responses, and recovery rather than summing potentially overlapping component durations. Tok accumulates input and completion tokens from every model call in an episode, including planning, reflection, recovery, and failed attempts, while excluding deterministic perception computation. Separately reported reasoning tokens, model-backed tools, and missing usage records must be identified rather than silently treated as ordinary tokens or zeros. The main verification, AndroidWorld, and runtime-ablation tables express Tok in thousands; dedicated appendix token tables retain unscaled counts. Relative differences are calculated from unscaled supplied values rather than rounded display values.

\subsection{Extended quantitative analysis}
\label{app:full_quantitative_analysis}
\label{app:full_token_results}

\subsubsection{Dynamic verification across backbones}

\paragraph{Success gains span different backbone--harness combinations.} Table~\ref{tab:hll_success_gaps} compares \frameworkname{} with the highest-success baseline for each backbone. UI-TARS Harness is the strongest baseline for GPT-5.4, MiniMax-M2.7, and MiniMax-M3; MobileAgent is strongest for the remaining backbones. \frameworkname{} improves upon both types of baseline across these configurations, indicating that its gains are not confined to a single baseline harness.

\begin{table}[!htbp]
    \centering
    \caption{Verification success relative to the highest-success
    baseline for each backbone. SR is in percent and $\Delta$ is a
    percentage-point difference calculated from the reported values.}
    \label{tab:hll_success_gaps}
    \small
    \setlength{\tabcolsep}{4pt}
    \begin{tabular}{@{}llrrr@{}}
        \toprule
        Backbone & Best baseline & Baseline SR & Our SR & $\Delta$ \\
        \midrule
        GPT-5.5
        & MobileAgent & 66.7 & 80.0 & +13.3 \\

        GPT-5.4
        & UI-TARS Harness & 15.6 & 46.7 & +31.1 \\

        Claude Opus 4.8
        & MobileAgent & 61.1 & 75.6 & +14.5 \\

        Claude Opus 4.6
        & MobileAgent & 38.9 & 34.4 & -4.5 \\

        \shortstack[l]{Gemini 3.1\\Pro Preview}
        & MobileAgent & 42.2 & 76.7 & +34.5 \\

        GLM-5v-turbo
        & MobileAgent & 38.9 & 57.8 & +18.9 \\

        MiniMax-M2.7
        & UI-TARS Harness & 10.0 & 32.2 & +22.2 \\

        MiniMax-M3
        & UI-TARS Harness & 12.2 & 43.3 & +31.1 \\

        Qwen3.7-plus
        & MobileAgent & 44.4 & 65.6 & +21.2 \\
        \bottomrule
    \end{tabular}
\end{table}

\paragraph{Relative improvements coexist with remaining completion failures.} MiniMax-M2.7 improves by 22.2 percentage points over its strongest baseline but reaches 32.2\% success, leaving substantial headroom. Claude Opus 4.6 exhibits a different trade-off: compared with MobileAgent, E2E decreases from 202.2 to 185.2 seconds while success decreases from 38.9\% to 34.4\%. This aggregate result motivates backbone-specific execution analysis rather than attributing the regression to a particular perceptual or reasoning weakness without trajectory evidence.

\paragraph{Generation volume alone does not explain latency gains.} Table~\ref{tab:hll_tokens} reports exact mean total-token counts. With GPT-5.5, \frameworkname{} uses 37,796 tokens versus MobileAgent's 23,817, yet achieves lower E2E. With Gemini 3.1 Pro Preview, its 59,443 tokens are below both MobileAgent's 63,973 and UI-TARS Harness's 92,021. Thus, the observed latency gains can coexist with either higher or lower generation volume. Token totals capture one component of execution cost; model-call scheduling and tool or environment waiting require their own trace-level measurements.

\begin{table}[t]
    \centering
    \caption{Mean total tokens in the dynamic verification evaluation.
    --- denotes an unavailable value.}
    \label{tab:hll_tokens}
    \small
    \setlength{\tabcolsep}{4pt}
    \begin{tabular}{@{}lrrrr@{}}
        \toprule
        Backbone & \frameworkname{}
        & MobileAgent & UI-TARS Harness & AppAgent \\
        \midrule
        GPT-5.5
        & 37,796 & 23,817 & 59,298 & 33,664 \\

        GPT-5.4
        & 74,934 & 37,369 & 73,006 & 56,344 \\

        Claude Opus 4.8
        & 77,539 & 52,721 & 81,543 & 81,663 \\

        Claude Opus 4.6
        & 47,756 & 30,754 & 77,375 & 23,434 \\

        \shortstack[l]{Gemini 3.1\\Pro Preview}
        & 59,443 & 63,973 & 92,021 & 26,268 \\

        GLM-5v-turbo
        & 97,493 & 42,802 & 77,924 & 38,366 \\

        MiniMax-M2.7
        & 68,728 & 28,285 & 22,120 & 8,968 \\

        MiniMax-M3
        & 135,543 & 76,076 & 80,482 & 43,461 \\

        Qwen3.7-plus
        & 56,985 & 67,488 & 84,423 & 31,928 \\
        \bottomrule
    \end{tabular}
\end{table}

\subsubsection{Application workflow execution}
\label{app:aw_extended_analysis}

\paragraph{Application-level gains include reduced latency and generation cost.} Against MobileAgent, \frameworkname{} improves success while reducing both E2E and total-token usage. The token reductions range from 41.86\% to 59.56\%, accompanying E2E reductions of 68.47\%--80.01\%. Table~\ref{tab:androidworld_tokens} provides the unscaled generation counts. These results support evaluating runtime design at the application-execution level as well as at verification gates.

\paragraph{Higher completion can require more processing than a lower-success system.} Compared with AppAgent, \frameworkname{} achieves higher success across the reported backbones and uses more tokens. E2E is lower with GPT-5.5, Gemini 3.1 Pro Preview, GLM-5v-turbo, and MiniMax-M3, but higher with Claude Opus 4.8 and Qwen3.7-Plus. For MiniMax-M3, the comparison is 211.74 versus 268.9 seconds. These differences show that improved completion and reduced generation volume are distinct outcomes. A lower all-run mean does not establish faster completion on identical successful subsets.

\paragraph{The UI-TARS harness is the strongest baseline on some backbones but not others.} With a general-purpose backbone, the UI-TARS observation--action procedure reaches 37.07\% success with GPT-5.5 and 31.03\% with MiniMax-M3, the second-best result for the latter backbone. Its success falls to 15.52\% with GLM-5v-turbo and 16.38\% with Qwen3.7-plus, where AppAgent and MobileAgent respectively rank second. Its E2E also varies widely, from 163.59 seconds with MiniMax-M3 to 475.13 seconds with Claude Opus 4.8. \frameworkname{} exceeds this baseline on every backbone in the comparison, by 8.67 to 38.78 percentage points. The ordering among baselines is therefore backbone-dependent, which is why the main comparison reports each harness separately rather than a single representative baseline.

\begin{table}[t]
    \centering
    \caption{Mean total tokens on AndroidWorld.
    --- denotes an unavailable result.}
    \label{tab:androidworld_tokens}
    \small
    \setlength{\tabcolsep}{4pt}
    \begin{tabular}{@{}lrrrr@{}}
        \toprule
        Backbone & \frameworkname{}
        & MobileAgent & UI-TARS Harness & AppAgent \\
        \midrule
        GPT-5.5
        & 100,855.07 & 173,468.95 & 29,383 & 45,736 \\

        Claude Opus 4.8
        & 135,585.61 & 256,178.60 & 44,583 & 29,169 \\

        \shortstack[l]{Gemini 3.1\\Pro Preview}
        & 83,671.19 & 160,938.70 & 23,539 & 29,724 \\

        GLM-5v-turbo
        & 134,372.59 & 310,902.90 & 63,440 & 53,478 \\

        MiniMax-M3
        & 136,867.59 & 269,488.01 & 30,968 & 15,142 \\

        Qwen3.7-plus
        & 92,594.98 & 228,954.11 & 31,353 & 40,159 \\
        \bottomrule
    \end{tabular}
\end{table}

\begin{figure}[htbp]
    \centering
    \includegraphics[width=0.9\linewidth]{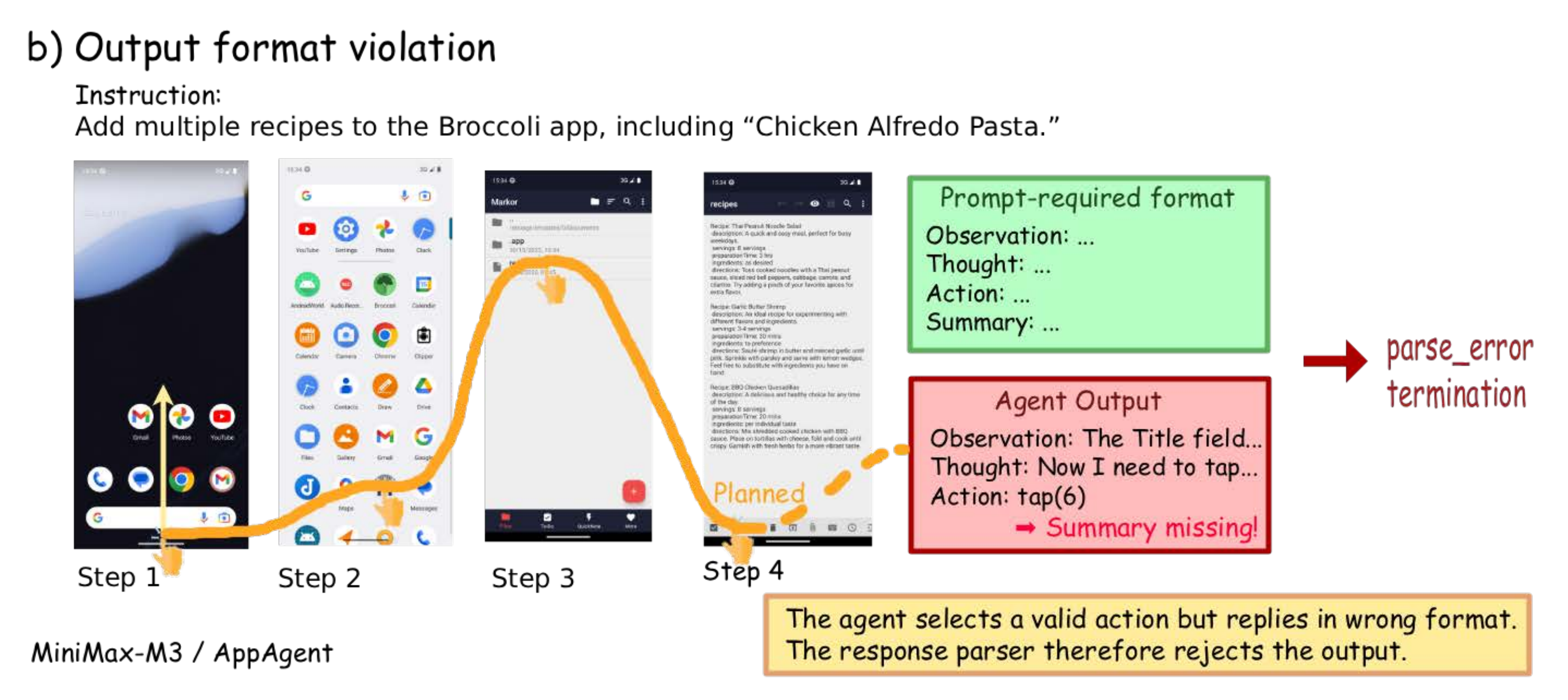}
    \caption{A-A2. Recipe entry with MiniMax-M3 / AppAgent. The agent selects a valid action but omits the required Summary field from its response; the parser rejects the output and the episode ends in a parse error.}
    \label{fig:aw2_appagent}
\end{figure}

\paragraph{Output-format failures warrant separate diagnosis.}\label{app:aw_failure_notes} AppAgent with MiniMax-M3 achieves 16.4\% success, 268.9 seconds of E2E, and 15,142 mean tokens. Its run notes report frequent ParseErrors caused by noncompliant output formats, illustrated by case A-A2 in Figure~\ref{fig:aw2_appagent}. This observation identifies an execution-interface failure mode, which is distinct from a task-level failure: the selected action was valid and the run ended at the response parser.

\subsection{Native UI-TARS evaluation}
\label{app:ui_tars_specialist}
\label{app:ui_tars_native_results}

\paragraph{The full evaluation confirms a substantial verification bottleneck.} UI-TARS-1.5-7B with its associated harness completes 325 of 900 verification tasks, achieving 36.11\% success. This 900-instance evaluation is separate from the 90-instance backbone--harness comparisons in Table~\ref{tab:hll_main}, which do not include this native policy. Table~\ref{tab:ui_tars_native_summary} also reports the AndroidWorld result of 23/116, or 19.83\% success, with 95.24 seconds of E2E and 14,018.81 tokens per run. Verification and application results are reported separately.

\begin{table}[t]
    \centering
    \caption{Native UI-TARS-1.5-7B with its associated harness.
    E2E is in seconds; Tok is the mean total-token count per run.
    Verification and application results use separate scored populations.}
    \label{tab:ui_tars_native_summary}
    \small
    \setlength{\tabcolsep}{5pt}
    \begin{tabular}{@{}lrrrr@{}}
        \toprule
        Setting & Success / Total & SR (\%) & E2E & Tok \\
        \midrule
        Dynamic verification
        & 325/900 & 36.11 & 116.38 & 19,909.43 \\

        AndroidWorld
        & 23/116 & 19.83 & 95.24 & 14,018.81 \\
        \bottomrule
    \end{tabular}
\end{table}

\paragraph{Success varies sharply across task families.} Table~\ref{tab:ui_tars_specialist} reports the per-family breakdown of the 900-instance evaluation, with 100 instances per family summing to the overall 325 successes. The distribution is highly uneven. Natural-image sequence selection reaches 98/100, and missing-patch selection 64/100, whereas board reconfiguration reaches 5/100, logic-and-arithmetic interaction 12/100, and category-guided image selection 13/100. Families whose targets are natural objects with distinct appearances are largely solved, while those requiring multi-step spatial manipulation or selection among visually similar candidates are not.

This ordering is consistent with the UI-TARS failure cases in Appendix~\ref{app:baseline_cases}. Slider alignment (22/100) and jigsaw alignment (38/100) both require a drag whose endpoint must satisfy a geometric criterion, matching the interaction-level difficulty in M-C2. The low result on logic-and-arithmetic interaction matches T-C1, where two candidates differing only by orientation were not separated. Icon sequence selection (49/100) sits between these groups, as in M-C1, where an otherwise valid ordered interaction failed at a single position. The aggregate 36.11\% therefore reflects a bimodal distribution rather than uniform partial competence.

\paragraph{Cost rises as success falls across families.} The per-family E2E and Tok values in Table~\ref{tab:ui_tars_specialist} are strongly and negatively associated with success: the Pearson correlation between family success count and mean total tokens is $-0.99$, and between success count and mean E2E is $-0.88$. The extremes illustrate the pattern. Natural-image sequence selection, solved in 98 of 100 instances, averages 33.87 seconds and 4,080 tokens. Board reconfiguration, solved in 5 instances, averages 138.66 seconds and 28,418 tokens, roughly seven times the generation cost for a twentieth of the success rate. Slider alignment records the longest mean runtime at 157.94 seconds with 22 successes.

The families the policy cannot solve are therefore also the ones on which it spends the most, which is the aggregate signature of repeated unproductive attempts rather than early abandonment. This is consistent with the recorded UI-TARS trajectories in Appendix~\ref{app:baseline_cases}: in T-C1 and T-C2 the agent continues to act on a target it cannot discriminate, and in M-C2 an invalid interaction form is retried while a different parameter is varied. Because the failing families consume disproportionate wall-clock time and generation, their cost is not recovered by terminating sooner. This is the same relationship the runtime ablations exhibit within \frameworkname{} (Section~\ref{sec:runtime_ablation}), where the configurations with lower success also incur higher E2E and token usage.

\begin{table}[t]
    \centering
    \caption{Per-family results for the full native UI-TARS verification evaluation. Each family contains 100 instances; successes sum to the overall total of 325, and the family means of E2E and Tok reproduce the overall values in Table~\ref{tab:ui_tars_native_summary}. Tok excludes image tokens.}
    \label{tab:ui_tars_specialist}
    \small
    \setlength{\tabcolsep}{5pt}
    \begin{tabular}{@{}lrrrr@{}}
        \toprule
        HLL task family & Instances & Successes & E2E (s) & Tok \\
        \midrule
        Natural-image sequence selection & 100 & 98 & 33.87 & 4,079.50 \\
        Icon sequence selection & 100 & 49 & 124.29 & 17,730.70 \\
        Slider alignment & 100 & 22 & 157.94 & 25,756.80 \\
        Jigsaw alignment & 100 & 38 & 140.64 & 19,441.90 \\
        Missing-patch selection & 100 & 64 & 68.55 & 12,027.70 \\
        Tile restoration & 100 & 24 & 123.28 & 22,689.10 \\
        Board reconfiguration & 100 & 5 & 138.66 & 28,418.00 \\
        Category-guided image selection & 100 & 13 & 130.39 & 23,604.70 \\
        Logic-and-arithmetic interaction & 100 & 12 & 129.76 & 25,436.50 \\
        \midrule
        Total / mean & 900 & 325 & 116.38 & 19,909.43 \\
        \bottomrule
    \end{tabular}
\end{table}

\subsection{Statistical Stability Analysis}
\label{app:repeated_run_results}

Repeated evaluation supplements the main comparisons by examining variation in success, latency, and generation cost. Table~\ref{tab:repeated_raw} reports mean and standard deviation over three repetitions for GPT-5.5, Claude Opus 4.8, and Gemini 3.1 Pro Preview. In all three backbones, \frameworkname{} attains the highest mean success rate and the lowest mean E2E among the four systems.

The separation in success exceeds the observed run-to-run dispersion in each case. With GPT-5.5, \frameworkname{} reaches $87.4 \pm 8.3$ against MobileAgent's $60.0 \pm 6.9$; with Claude Opus 4.8, $78.15 \pm 3.9$ against $58.5 \pm 3.6$; with Gemini 3.1 Pro Preview, $71.85 \pm 7.0$ against $36.3 \pm 6.5$. The latency gaps are larger still relative to their dispersion, most clearly with Gemini 3.1 Pro Preview, where 130.9 $\pm$ 19.9 seconds compares against MobileAgent's 514.4 $\pm$ 174.8. The baselines' own dispersion is uneven: MobileAgent's E2E standard deviation reaches 174.8 seconds with Gemini 3.1 Pro Preview and 82.5 with Claude Opus 4.8, whereas \frameworkname{} stays between 12.6 and 19.9 seconds across the three backbones, indicating more consistent execution time as well as a lower mean.

The three repetitions use seeds 42, 43, and 45, with 90 tasks per run (10 from each of the nine families). The task instances are fixed across repetitions and match the main evaluation instance set; the difficulty policy and distraction level are also unchanged. Mean and standard deviation describe run-to-run variation on this fixed set. They summarize three runs rather than the single run reported in Table~\ref{tab:hll_main}, so their means need not equal the main-table entries. We report this dispersion descriptively and do not infer significance from the three aggregate run summaries.

\begin{table}[t]
    \centering
    \caption{Repeated-evaluation summaries over three runs, reported as
    mean $\pm$ standard deviation.}
    \label{tab:repeated_raw}
    \small
    \setlength{\tabcolsep}{4pt}
    \begin{tabular}{@{}lrrr@{}}
        \toprule
        Framework & SR (\%) & E2E (s) & Tok \\
        \midrule
        \multicolumn{4}{l}{\textit{GPT-5.5}} \\
        \frameworkname{}
        & \textbf{87.4} $\pm$ 8.3 & \textbf{98.6} $\pm$ 19.6 & 40,921 $\pm$ 3,438 \\

        MobileAgent
        & 60.0 $\pm$ 6.9 & 206.9 $\pm$ 6.6 & 25,937 $\pm$ 518 \\

        UI-TARS Harness
        & 41.9 $\pm$ 1.3 & 204.3 $\pm$ 28.4 & 58,621 $\pm$ 1,330 \\

        AppAgent
        & 36.30 $\pm$ 3.39 & 209.69 $\pm$ 26.28
        & 30,085.74 $\pm$ 3,266.41 \\

        \midrule
        \multicolumn{4}{l}{\textit{Claude Opus 4.8}} \\
        \frameworkname{}
        & \textbf{78.15} $\pm$ 3.9 & \textbf{76.5} $\pm$ 12.6 & 82,183 $\pm$ 6,568 \\

        MobileAgent
        & 58.5 $\pm$ 3.6 & 319.0 $\pm$ 82.5 & 53,509 $\pm$ 16,646 \\

        UI-TARS Harness
        & 31.9 $\pm$ 7.4 & 181.1 $\pm$ 47.3 & 105,892 $\pm$ 10,503 \\

        AppAgent
        & 11.9 $\pm$ 6.4 & 255.6 $\pm$ 20.3 & 54,195 $\pm$ 10,217 \\

        \midrule
        \multicolumn{4}{l}{\textit{Gemini 3.1 Pro Preview}} \\
        \frameworkname{}
        & \textbf{71.85} $\pm$ 7.0 & \textbf{130.9} $\pm$ 19.9 & 59,908 $\pm$ 659 \\

        MobileAgent
        & 36.3 $\pm$ 6.5 & 514.4 $\pm$ 174.8 & 48,637 $\pm$ 16,094 \\

        UI-TARS Harness
        & 13.7 $\pm$ 2.3 & 378.0 $\pm$ 48.7 & 86,509 $\pm$ 5,196 \\

        AppAgent
        & 31.9 $\pm$ 3.6 & 275.2 $\pm$ 41.3 & 30,758 $\pm$ 1,675 \\
        \bottomrule
    \end{tabular}
\end{table}

\subsection{Paired analysis of static correctness and dynamic acceptance}
\label{app:paired_acceptance}

We examine 90 paired HLL tasks with GPT-5.5 and compare MobileAgent with \frameworkname{}. Static answer correctness is determined by the static checker, while dynamic acceptance is determined by the dynamic validator. The targeted failure class requires an actually submitted correct answer, dynamic rejection, and final episode failure. Merely identifying part of a solution or failing to submit is not sufficient for this class. The same task is then matched to the \frameworkname{} outcome.

\begin{table}[!htbp]
    \centering
    \caption{Paired diagnostic counts for GPT-5.5 on 90 HLL tasks. Conditional success refers to the baseline-defined failure subset, not a causal repair rate.}
    \label{tab:paired_acceptance}
    \small
    \begin{tabularx}{\linewidth}{@{}Xrr@{}}
        \toprule
        Quantity & Count & Proportion \\
        \midrule
        MobileAgent: static-correct, dynamically rejected, final failure & 13/90 & 14.44\% \\
        \frameworkname{} success on those same 13 tasks & 7/13 & 53.85\% \\
        MobileAgent failure, \frameworkname{} success & 13/90 & 14.44\% \\
        MobileAgent success, \frameworkname{} failure & 1/90 & 1.11\% \\
        \bottomrule
    \end{tabularx}
\end{table}

The paired success difference is $(13-1)/90=13.33$ percentage points. The reverse conversion is an incorrect task solution rather than a dynamically rejected correct answer. These counts distinguish the existence of correct-answer rejection from the overall system improvement on matched tasks. They do not establish that the remaining six tasks in the baseline-defined subset are all dynamic rejections under \frameworkname{}, or that binding alone explains their outcomes. Tool and execution-interface differences remain part of the system comparison; the binding-only ablation provides a separate controlled test.

\subsection{Mechanism-Level Trace Statistics}
\label{app:mechanism_trace_statistics}

This section reports per-step decision-log counts for \frameworkname{} with GPT-5.5, Claude Opus 4.8, and Gemini 3.1 Pro Preview. Each logged batch contains 90 episodes (nine HLL families, 10 instances each), using distraction level~2 and \texttt{hard} difficulty where supported, otherwise \texttt{normal}. These logs provide a separate mechanism analysis from the aggregate main table and the 225-episode ablation; the difficulty policy is shared. Counts are obtained by offline analysis without additional model calls.

For each step with a recorded decision, the supervisor's resolution is logged as \texttt{accepted} (the proposal was dispatched as given) or \texttt{overridden} (the dispatched action differed from the model's proposal). Table~\ref{tab:mechanism_trace} reports the number of decision steps with a recorded resolution, the count and rate of overridden steps, and the number of steps at which a bound action was recorded as supporting the decision (\texttt{canonical\_action} in the log schema).

\begin{table}[t]
    \centering
    \caption{Per-step decision-log counts for \frameworkname{}, from a 90-episode supplementary batch per backbone (\texttt{hard} where supported, otherwise \texttt{normal}; distraction level~2). ``Override rate'' is the fraction of decision steps with a recorded resolution that were overridden. This batch is not the 90-instance main table or the 225-episode ablation.}
    \label{tab:mechanism_trace}
    \small
    \setlength{\tabcolsep}{5pt}
    \begin{tabular}{@{}lrrr@{}}
        \toprule
        & GPT-5.5 & Claude Opus 4.8 & Gemini 3.1 Pro \\
        \midrule
        Saved steps & 378 & 585 & 621 \\
        Steps with resolution & 376 & 585 & 615 \\
        Steps without resolution & 2 & 0 & 6 \\
        Overridden steps & 26 & 78 & 79 \\
        Override rate & 6.91\% & 13.33\% & 12.85\% \\
        Overrides per episode & 0.289 & 0.867 & 0.878 \\
        Steps with a bound action & 82 & 104 & 120 \\
        \bottomrule
    \end{tabular}
\end{table}
\FloatBarrier

\paragraph{Override categories distinguish mapping from control.} We assign each overridden step one primary category, so category counts sum to the override totals (Table~\ref{tab:override_categories}). Equivalent action-name or parameter normalization accounts for 4, 15, and 18 steps; commitment-based changes of the actual operation account for 7, 28, and 26. Readiness-triggered automatic submission accounts for 11, 16, and 8, and budget-triggered control for 4, 19, and 27. An override therefore does not necessarily correct a mistaken proposal. Likewise, a bound-action step records supporting state at a decision step, not a distinct commitment or necessarily an override.

The distributions differ across backbones. Automatic submission is the largest category for GPT-5.5, commitment-based operation changes for Claude Opus 4.8, and budget-triggered control for Gemini 3.1 Pro Preview. Across the three batches, 61 of 183 overrides change the operation according to a commitment, while 37 normalize equivalent operations. These are frequencies of control paths, not estimates of corrective success. A higher override rate does not by itself imply more accurate supervision.

\begin{table}[!htbp]
    \centering
    \caption{Primary category of overridden decision steps. Each step is counted once; counts are not episode-level success rates.}
    \label{tab:override_categories}
    \small
    \begin{tabularx}{\linewidth}{@{}Xrrr@{}}
        \toprule
        Category & GPT-5.5 & Opus 4.8 & Gemini 3.1 Pro \\
        \midrule
        Equivalent name/parameter normalization & 4 & 15 & 18 \\
        Commitment-based operation change & 7 & 28 & 26 \\
        Readiness-triggered automatic submission & 11 & 16 & 8 \\
        Budget-triggered control & 4 & 19 & 27 \\
        \midrule
        Total overrides & 26 & 78 & 79 \\
        \bottomrule
    \end{tabularx}
\end{table}
\FloatBarrier

\subsection{Runtime ablations and processing-cost analysis}
\label{app:runtime_ablation_details}

\subsubsection{Shared settings and control changes}
\label{app:ablation_controls}

\paragraph{Shared settings.} All six configurations use GPT-5.5, the same 225 HLL instances, public tools, model-visible evidence, task instructions, execution interfaces, outer time and action limits, and evaluator permissions. The no-binding condition only removes the forced dispatch precedence of evidence-supported pending operations; it retains automatic submission, readiness checks, evidence recording, model access to that evidence, legality checks, recovery, and all other controls. Relaxed supervision jointly removes forced binding and automatic submission. No condition receives answers or additional evaluator information.

\begin{table}[t]
    \centering
    \caption{Runtime-control configurations used in the ablation. Each variant modifies a specific control mechanism while retaining the execution paths listed in the final column.}
    \label{tab:runtime_ablation_configs}
    \footnotesize
    \setlength{\tabcolsep}{5pt}
    \renewcommand{\arraystretch}{1.16}
    \begin{tabularx}{\linewidth}{@{}>{\raggedright\arraybackslash}p{0.19\linewidth}>{\raggedright\arraybackslash}X>{\raggedright\arraybackslash}X@{}}
        \toprule
        \textbf{Configuration} & \textbf{Control change} & \textbf{Retained execution paths} \\
        \midrule
        \rowcolor{baselinebg}
        \textbf{Full supervision} & Evidence-supported pending operations take dispatch precedence. Submission readiness and budget state may trigger submission, checking, refresh, recovery, or termination. & Parsing, validation, execution, state updates, public feedback, budget enforcement, and terminal evaluation. \\
        \addlinespace[3pt]
        \cellcolor{configbg}\textbf{Without complete mediation} & Retained state and feedback no longer constrain the dispatched proposal through supervisory replacement. & Parsing and schema checks, executor validation, outer budgets, state recording, public feedback, and terminal evaluation. \\
        \addlinespace[3pt]
        \cellcolor{configbg}\textbf{Without fail-safe defaults} & Primary mediation, pending-operation binding, and readiness-based submission remain, while conservative submission restrictions and budget-triggered safeguards are relaxed. & Outer budgets, tool execution, page verification, state recording, and terminal evaluation. \\
        \addlinespace[3pt]
        \cellcolor{configbg}\textbf{No binding} & Evidence-supported pending operations no longer have forced dispatch precedence over model proposals. & Model-visible evidence, parsing, legality checks, executor, automatic submission and readiness checks, recovery, and budgets. \\
        \addlinespace[3pt]
        \cellcolor{configbg}\textbf{Relaxed supervision} & Pending operations are no longer forcibly dispatched, and readiness alone no longer forces submission. & Recovery state, associated budgets, outer limits, alternative submission paths, and terminal evaluation. \\
        \addlinespace[3pt]
        \cellcolor{configbg}\textbf{Unbounded recovery} & The per-category cap on recovery budgets ($b_k^{(j)}$ in Appendix~\ref{app:recovery_details}) is removed; recovery can continue until the outer episode timeout, without otherwise changing mediation, defaults, or submission logic. & Mediation, pending-operation binding, fail-safe defaults, readiness-based submission, and terminal evaluation. \\
        \bottomrule
    \end{tabularx}
\end{table}
\FloatBarrier

\subsubsection{Success--cost analysis}

\paragraph{Success and cost comparisons.} The six configurations record 185, 164, 169, 161, 154, and 188 successes out of 225 for full control, no complete mediation, no fail-safe defaults, no binding, relaxed supervision, and unbounded recovery, respectively. SR differences are calculated from these counts before rounding. In particular, $3/225$ is 1.33 percentage points, although subtracting the displayed rounded SR values yields 1.34. Table~\ref{tab:ablation_deltas} reports the differences. For each ablated configuration $a$, cost reduction is computed as
\begin{equation}
    R_C(a)=\frac{C(a)-C(\mathrm{full})}{C(a)},
    \qquad C\in\{\mathrm{E2E},\mathrm{Tok}\}.
    \label{eq:ablation_cost_reduction}
\end{equation}
The denominator is the cost of the ablated configuration. Success improvements are reported as percentage-point differences; for unbounded recovery this difference is negative, since that configuration attains slightly higher success than full control.

\begin{table}[t]
    \centering
    \caption{Full control relative to each ablation. Success gains are
    in percentage points; cost reductions use the ablated condition
    as the denominator. A negative success gain means the ablated
    configuration attains higher success than full control.}
    \label{tab:ablation_deltas}
    \small
    \setlength{\tabcolsep}{4pt}
    \begin{tabular}{@{}lrrr@{}}
        \toprule
        Ablated configuration
        & SR gain & E2E reduction & Tok reduction \\
        \midrule
        Without complete mediation
        & +9.33 & 9.51\% & 13.04\% \\

        Without fail-safe defaults
        & +7.11 & 29.09\% & 31.03\% \\

        No binding
        & +10.67 & 12.88\% & 15.21\% \\

        Relaxed supervision
        & +13.78 & 6.53\% & 10.71\% \\

        Unbounded recovery
        & $-$1.33 & 73.14\% & 58.76\% \\
        \bottomrule
    \end{tabular}
\end{table}

\paragraph{Bypassing mediation does not reduce overall execution cost.} Without complete mediation, success falls by 9.33 percentage points, mean E2E increases by 11.79 seconds, and mean output grows by 6,273 tokens. The runtime layer's net value therefore exceeds its additional control work in this evaluation. The result is consistent with the intended role of mediation in carrying evidence-supported operations into execution. Attribution to particular proposal replacements or recovery paths requires the corresponding dispatch records.

\paragraph{Removing default controls creates the largest cost increase.} The fail-safe-defaults ablation adds 46.00 seconds and 18,818 tokens per run, the largest increases among the mediation, defaults, binding, and relaxed-supervision variants, while reducing success by 7.11 points. Removing controls can therefore increase total processing rather than make execution cheaper. One hypothesis for trace-level examination is that permitting some requests without sufficient support increases subsequent correction or recovery. The aggregate measurements establish the net difference, while logs are needed to identify the responsible steps.

\paragraph{Binding contributes beyond automatic submission.} With automatic submission retained, removing binding reduces success from 185 to 161 episodes, while mean E2E increases by 16.58 seconds and mean tokens rise from 41.82k to 49.323k. Removing automatic submission as well yields 154 successes, seven fewer than the no-binding configuration. These comparisons estimate binding's contribution when automatic submission is enabled and automatic submission's contribution when binding is relaxed. They do not identify either component's effect in every setting or their interaction without the fourth factorial condition.

\paragraph{Recovery caps impose a success--cost trade-off.} Removing the caps increases success from 185 to 188 episodes, a net gain of three, while E2E rises from 112.13 to 417.45 seconds and tokens from 41.82k to 101.40k. Extra recovery thus retains an observed completion benefit at substantially greater cost. The aggregate difference does not reveal how many individual tasks improve versus regress, and cannot establish that attempts beyond the cap are uninformative.

\subsubsection{Connection to runtime design principles}

\paragraph{Separate controls answer different questions.} The no-binding condition isolates dispatch precedence while retaining automatic submission. The recovery-cap condition isolates the configured recovery limit while retaining binding and submission. The former supports enforcing available evidence in the tested configuration; the latter characterizes the success--cost trade-off of bounded recovery. Neither establishes that the runtime computes the optimal value of further information.

\paragraph{Relation to metareasoning.} Full control outperforms the mediation, defaults, binding, and relaxed-supervision variants in success and aggregate cost. Unbounded recovery instead slightly improves success at higher cost. This pattern motivates evaluating additional processing by its contribution to task completion relative to its expense. It supports the practical control design, without demonstrating an optimal metalevel policy or a rule that reliably detects when further information has no value.

\paragraph{Execution traces would localize where cost is incurred.} The aggregate measurements in Table~\ref{tab:runtime_ablation_main} establish net differences between configurations but do not identify which steps account for the additional latency and tokens of the ablated configurations. Per-episode counts of model calls, tool calls, dispatched GUI actions, commitment dispatches, proposal replacements, and recovery attempts would connect these configuration-level results to the mechanisms illustrated in Appendices~\ref{app:baseline_cases} and~\ref{app:solvebind_cases}. We do not infer such counts from the aggregate means, since several distinct step profiles are consistent with the same episode average.

\FloatBarrier
\subsection{Failure Mode Analysis}
\label{app:experimental_cases}
\label{app:baseline_cases}
\label{app:failure_taxonomy}

The analysis is organized around twelve baseline failure cases: four each for MobileAgent, UI-TARS, and AppAgent, consisting of two CAPTCHA tasks and two AndroidWorld tasks per baseline. Each case records the goal, critical operations, feedback, failure point, backbone, and termination status. UI-TARS cases specify whether they use a general-purpose backbone or UI-TARS-1.5-7B; all four UI-TARS cases reported here use general-purpose backbones with the UI-TARS harness. Table~\ref{tab:failure_modes} summarizes the cases and the mechanism each one isolates.

\begin{table}[!htbp]
    \centering
    \caption{Summary of the twelve baseline failure cases. Cases are described from recorded trajectories, with accompanying figures where the displayed material matches the task. Labels name the mechanism the case isolates; they are not population failure frequencies.}
    \label{tab:failure_modes}
    \footnotesize
    \setlength{\tabcolsep}{4pt}
    \renewcommand{\arraystretch}{1.15}
    \begin{tabularx}{\linewidth}{@{}>{\raggedright\arraybackslash}p{0.06\linewidth}>{\raggedright\arraybackslash}p{0.16\linewidth}>{\raggedright\arraybackslash}p{0.20\linewidth}>{\raggedright\arraybackslash}X@{}}
        \toprule
        \textbf{Case} & \textbf{Backbone} & \textbf{Termination} & \textbf{Mechanism isolated} \\
        \midrule
        \cellcolor{labelbg}\textbf{M-C1} & GPT-5.5 & Verification rejected & Grounding error at one position of an ordered sequence \\
        \cellcolor{labelbg}\textbf{M-C2} & Claude Opus 4.6 & 15 steps, no attempt & Retries vary the wrong parameter; interaction form invalid \\
        \cellcolor{labelbg}\textbf{M-A1} & GPT-5.5 & Step limit, reward 0 & Repeated correction without checking its effect \\
        \cellcolor{labelbg}\textbf{M-A2} & GPT-5.5 & Agent stop, reward 0 & Completion claimed from dispatch, not from evidence \\
        \addlinespace[2pt]
        \cellcolor{labelbg}\textbf{T-C1} & GPT-5.5 & Verification rejected & Candidates differing only by orientation not separated \\
        \cellcolor{labelbg}\textbf{T-C2} & GPT-5.5 & Verification rejected & Named target not bound to the region displaying it \\
        \cellcolor{labelbg}\textbf{T-A1} & Claude Opus 4.8 & Filename not set & Execution primitive drops non-alphabetic characters \\
        \cellcolor{labelbg}\textbf{T-A2} & Gemini 3.1 Pro & Goal not reached & Task target not retained over a longer sequence \\
        \addlinespace[2pt]
        \cellcolor{labelbg}\textbf{A-C1} & Gemini 3.1 Pro & False completion & Correct diagnosis stated, same action reissued \\
        \cellcolor{labelbg}\textbf{A-C2} & Claude Opus 4.8 & Verification rejected & Correct target named, grounded to wrong element \\
        \cellcolor{labelbg}\textbf{A-A1} & GLM-5v-turbo & Max-step & Wrong action primitive for the required effect \\
        \cellcolor{labelbg}\textbf{A-A2} & MiniMax-M3 & Parse error & Valid action rejected at the output-format interface \\
        \bottomrule
    \end{tabularx}
\end{table}
\FloatBarrier

\paragraph{Failure attribution follows observable execution evidence.} For each case, the analysis proceeds from the initial context to the critical proposal, dispatched operation, resulting feedback, and final outcome. Repeated solving is assessed against whether relevant new evidence was obtained. Premature progression is assessed against the state required by the next operation. Reused grounding is checked against target and page changes. Labels such as redundant solving, stale evidence reuse, retries without hypothesis revision, or output-format failure are assigned only when supported by the trace. The selected cases explain mechanisms rather than estimate population failure frequencies.

\providecommand{\SBCaseFigure}[3]{%
    \begin{figure}[htbp]
        \centering
        \IfFileExists{#1}{%
            \includegraphics[
                width=\linewidth,
                height=0.58\textheight,
                keepaspectratio
            ]{#1}%
        }{%
            \fbox{%
                \parbox[c][0.14\textheight][c]{0.90\linewidth}{%
                    \centering
                    Measured case figure pending\\[4pt]
                    \footnotesize
                    \texttt{\detokenize{#1}}
                }%
            }%
        }
        \caption{#3}
        \label{#2}
    \end{figure}%
    \FloatBarrier
}

\subsubsection{MobileAgent failure cases}

\paragraph{M-C1: icon-sequence selection (GPT-5.5 / MobileAgent).} The task requires clicking icons in a prescribed order. The agent reads the sequence correctly and executes the first two selections as required. At the third position it commits to a different icon from the one shown in the instruction strip, having reasoned that the required target was the scissors in the lower-left region, and then submits the sequence. The environment reports that the third click was not on the correct icon. Across the episode the run issues three verification attempts and records no successful verification. The grounding error is confined to a single position, but because acceptance depends on the whole ordered sequence, a locally correct prefix does not yield partial credit.

\begin{figure}[htbp]
    \centering
    \includegraphics[width=0.9\linewidth]{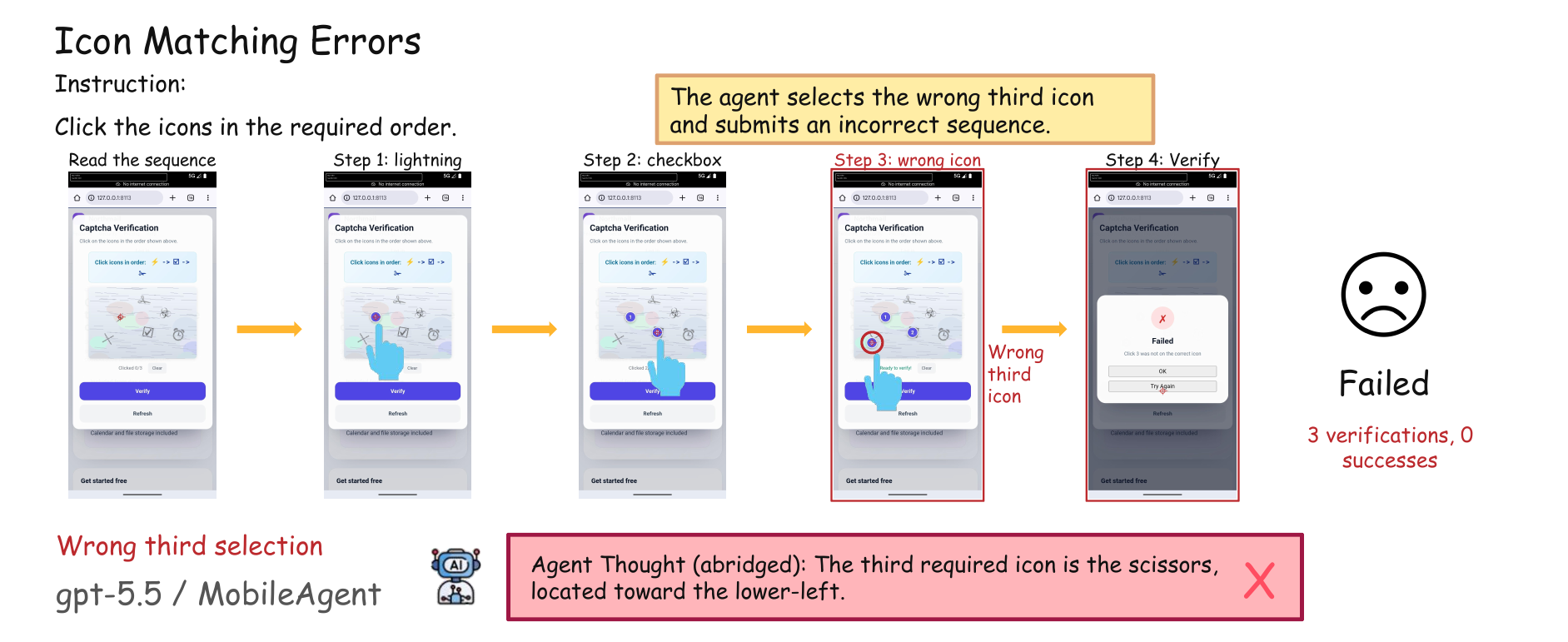}
    \caption{M-C1. Icon-sequence selection with GPT-5.5 / MobileAgent. The first two selections match the required order; the third resolves to the wrong icon, and verification is rejected. Three verification attempts produce no success.}
    \label{fig:ca1_mobileagent}
\end{figure}

\paragraph{M-C2: slider alignment (Claude Opus 4.6 / MobileAgent).} The task requires dragging a slider until the puzzle piece meets the gap. The agent begins the drag on the image area rather than on the slider handle, so the gesture does not move the piece and the page state is unchanged. Subsequent attempts vary the drag endpoint while repeating the same incorrect start point, with the recorded reasoning proposing a new $x$ coordinate not yet tried. The run consumes 15 steps and reaches no verification attempt. The unresolved condition concerns where the interaction must originate, whereas the retries vary a different parameter, so repeated attempts add execution cost without addressing the failure. This case differs from M-C1, where grounding resolved to a wrong target within an otherwise valid interaction; here the interaction form itself is invalid and no attempt reaches the verifier.

\begin{figure}[htbp]
    \centering
    \includegraphics[width=0.9\linewidth]{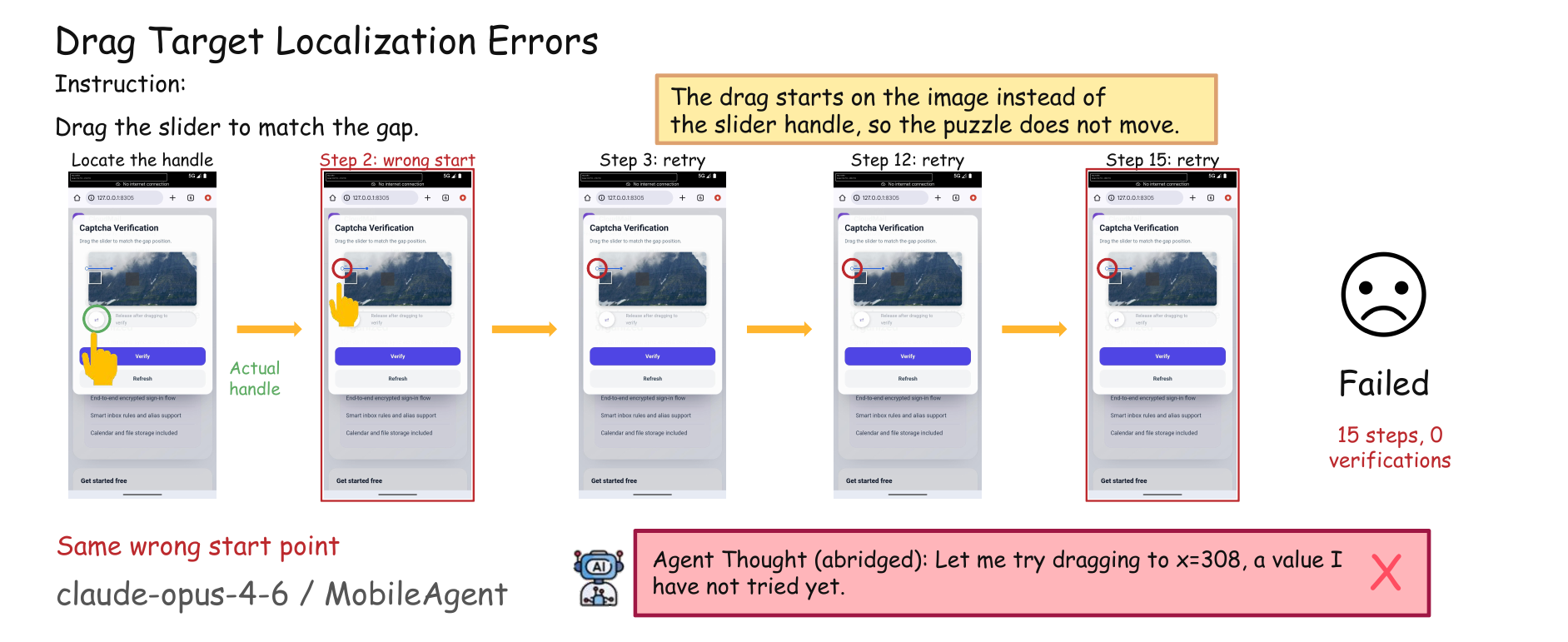}
    \caption{M-C2. Slider alignment with Claude Opus 4.6 / MobileAgent. Each drag starts on the image rather than on the handle (circled), so the piece never moves. Retries at steps 3, 12, and 15 repeat the same start point while varying the endpoint; 15 steps yield no verification attempt.}
    \label{fig:ca2_mobileagent}
\end{figure}

\paragraph{M-A1: audio file naming (GPT-5.5 / MobileAgent).} The task requires recording audio and saving it under a specified filename. The agent reaches the rename dialog and enters a name that does not match the target. It then recognizes that the filename is incorrect and attempts to clear the field, issuing 13 identical backspace taps across subsequent steps. The field content remains unchanged throughout, and the episode ends at the 20-step limit with zero reward. The agent's assessment of the state is correct, and the chosen recovery direction is appropriate; what fails is the check on whether the corrective action had any effect. Repetition continues because the absence of change is not itself treated as evidence against the current approach.

\begin{figure}[htbp]
    \centering
    \includegraphics[width=0.9\linewidth]{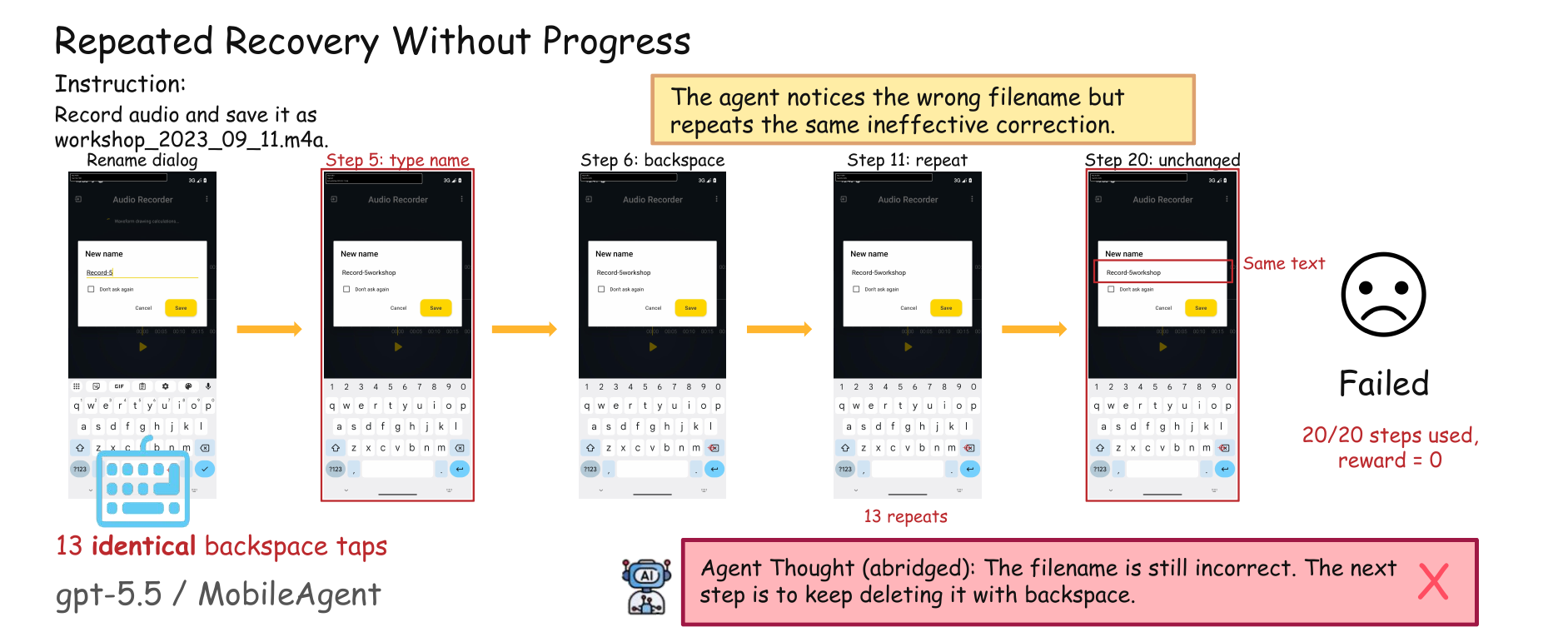}
    \caption{M-A1. Audio file naming with GPT-5.5 / MobileAgent. The agent detects the wrong filename and issues 13 identical backspace taps; the field text is unchanged between steps 6 and 20, and the run terminates at the step limit with zero reward.}
    \label{fig:aw1_mobileagent}
\end{figure}

\paragraph{M-A2: brightness setting (GPT-5.5 / MobileAgent).} The task requires setting display brightness to maximum. The agent opens the notification shade and expands it, then performs a left-to-right swipe across the region where the brightness slider would appear. The captured screen at that step shows no visible slider. The agent nevertheless records that the swipe should have set brightness to maximum and issues an explicit stop at the next step. The final state does not satisfy the task and the run receives zero reward. The failure is a transition from a dispatched action to a completion claim without evidence that the action took effect. Unlike M-A1, where the agent kept acting on an unchanged state, here it stops early on an assumed one; both outcomes follow from treating dispatch as progress.

\begin{figure}[htbp]
    \centering
    \includegraphics[width=0.9\linewidth]{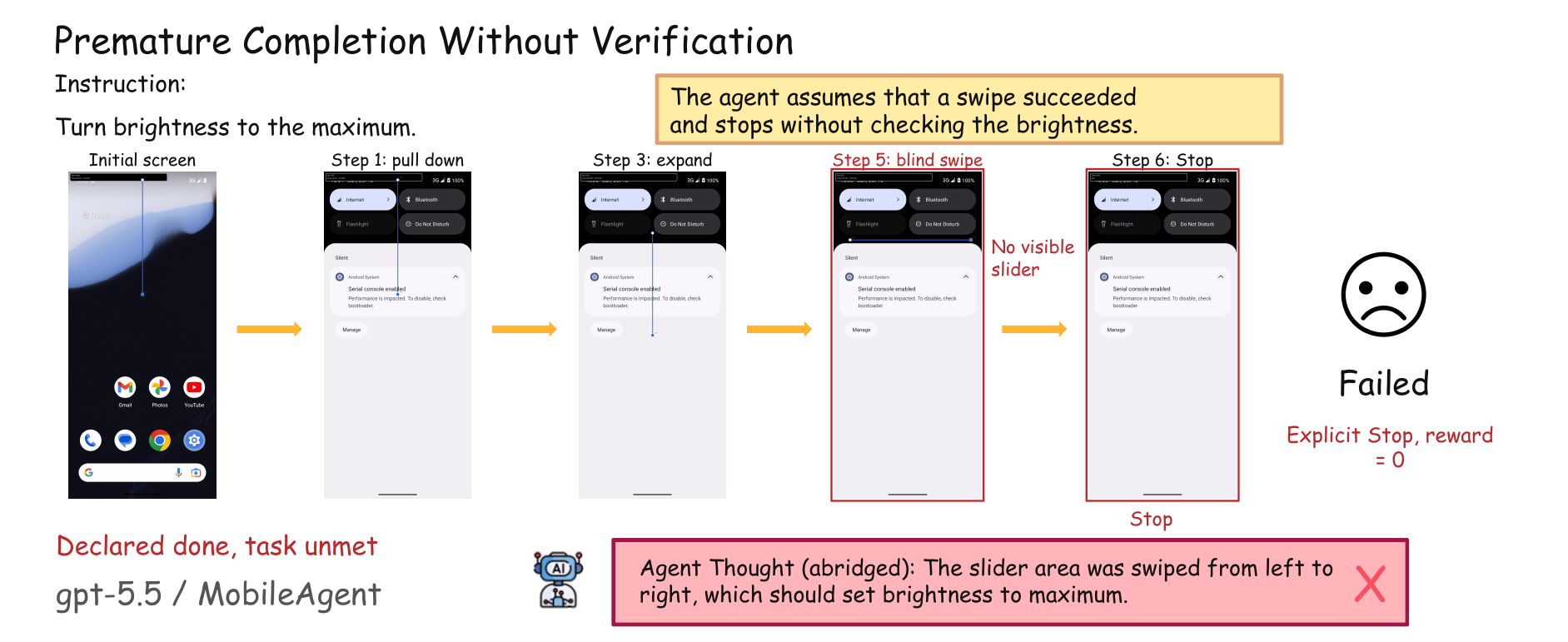}
    \caption{M-A2. Brightness setting with GPT-5.5 / MobileAgent. The swipe at step 5 targets a region where no slider is visible; the agent reports the action as successful and stops at step 6, leaving the task unsatisfied.}
    \label{fig:aw2_mobileagent}
\end{figure}

\subsubsection{UI-TARS failure cases}

\paragraph{T-C1: logic-and-arithmetic interaction (GPT-5.5 with the UI-TARS harness).} The task requires clicking in a scene so that a round marker lands on the uppercase letter V that is clearly rotated sideways. The scene contains two instances of the same letter that differ only in orientation. The agent states the requirement correctly, reporting that it will select the sideways one, but its clicks do not separate the two instances by rotation, and verification is rejected. The task content is understood at the level of the instruction; what the interaction does not supply is a grounding decision that distinguishes candidates differing only in a geometric attribute. This configuration uses a general-purpose backbone with the UI-TARS observation--action procedure, not the native UI-TARS-1.5-7B policy.

\begin{figure}[htbp]
    \centering
    \includegraphics[width=0.9\linewidth]{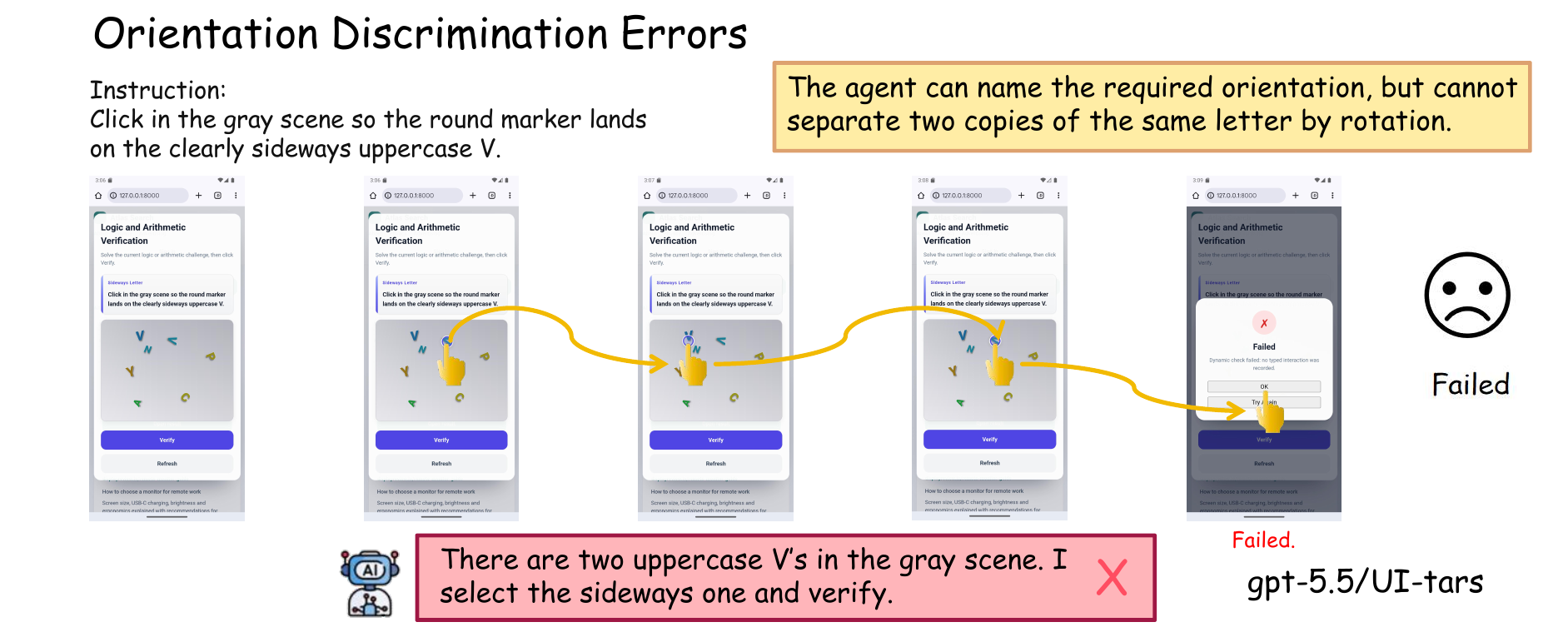}
    \caption{T-C1. Logic-and-arithmetic interaction with GPT-5.5 and the UI-TARS harness. Two copies of the same uppercase letter differ only by rotation; the agent names the required orientation but its selection does not separate them, and the dynamic check is rejected.}
    \label{fig:ca1_uitars}
\end{figure}

\paragraph{T-C2: scene character selection (GPT-5.5 with the UI-TARS harness).} The task requires clicking characters embedded in a photograph in the order V, C, W. The agent reads the required order correctly and clicks two markers in sequence, but the letters it selects are not the ones the corresponding circles actually display. The environment reports that the first click was not on the correct character. The failure concerns binding a named target to the region that displays it, rather than reading the instruction or ordering the steps. Together with T-C1, these cases locate the UI-TARS harness's verification bottleneck at fine-grained visual grounding inside cluttered scenes, in both cases under conditions where the required sequence was itself identified correctly.

\begin{figure}[htbp]
    \centering
    \includegraphics[width=0.9\linewidth]{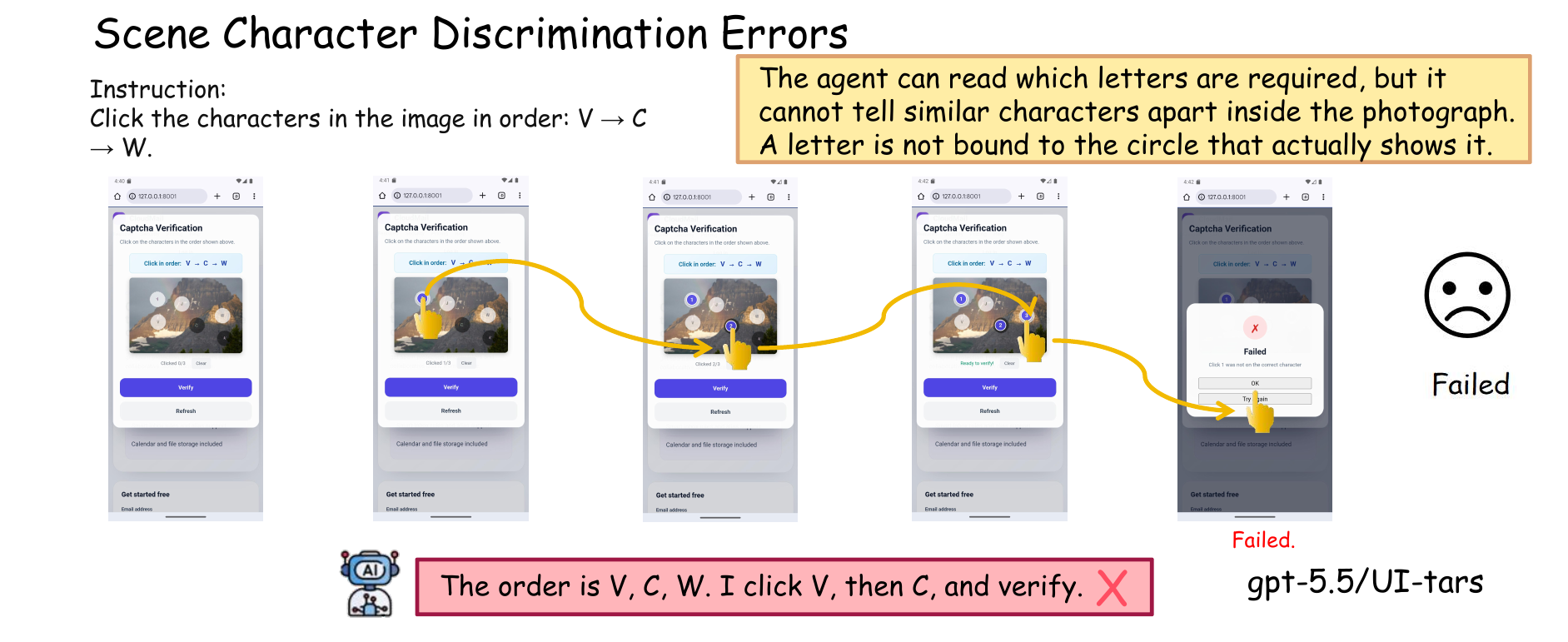}
    \caption{T-C2. Scene character selection with GPT-5.5 and the UI-TARS harness. The required order V, C, W is read correctly, but the selected circles do not contain the named characters; verification reports that the first click was incorrect.}
    \label{fig:ca2_uitars}
\end{figure}

\paragraph{T-A1: symbol-level text entry (Claude Opus 4.8 with the UI-TARS harness).} The task requires saving a recording under a filename containing an underscore and a file extension. The agent knows the target filename and types it in one action, but the field retains only the alphabetic prefix; the underscore and extension do not register. It then switches to the symbol keyboard to enter the missing character. The episode ends without the required filename. The failure is at the level of action execution rather than planning: the intended text is correct, while the soft-keyboard interaction does not reproduce non-alphabetic characters reliably. This mode is visible only because the agent checked the field contents after typing, which distinguishes it from M-A1, where an unchanged field went unexamined.

\begin{figure}[htbp]
    \centering
    \includegraphics[width=0.9\linewidth]{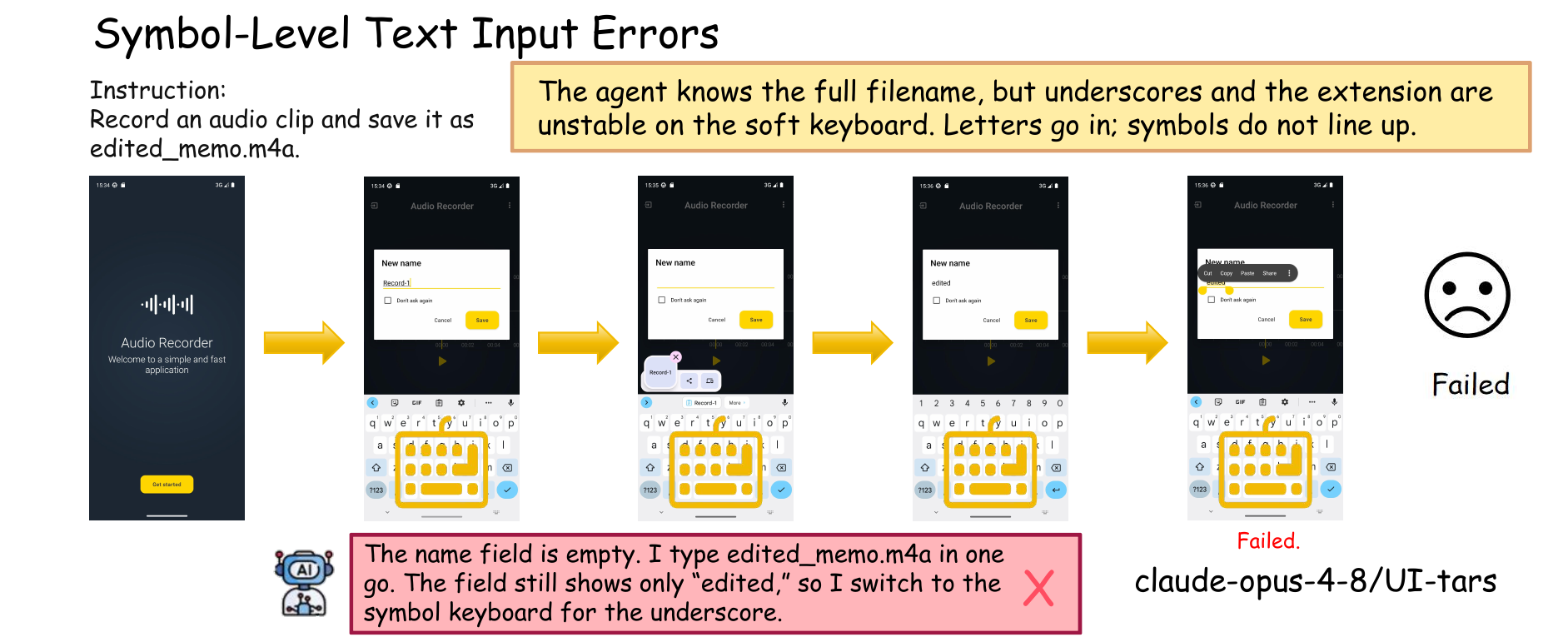}
    \caption{T-A1. Symbol-level text entry with Claude Opus 4.8 and the UI-TARS harness. The full filename is typed in one action, but the field shows only the alphabetic prefix; the underscore and extension are not entered, and the episode ends without the required name.}
    \label{fig:aw1_uitars}
\end{figure}

\paragraph{T-A2: long-horizon app workflow (Gemini 3.1 Pro Preview with the UI-TARS harness).} The task requires reading recipes from an image and adding them to a target application. After a few steps the agent no longer operates on the task application: the recorded screens move from the launcher through an audio recorder, a contacts setup page, and the notification shade, returning toward the home screen rather than progressing toward the goal. The run ends without the recipes being added. Whereas T-A1 fails at a single execution primitive within the correct context, here the difficulty is retaining the task target across a longer sequence, so later actions are no longer directed at the goal state. We note that the reasoning excerpt reproduced in the figure describes a slider interaction and does not correspond to this task; the step-level screens, not that excerpt, are the basis for this description.

\begin{figure}[htbp]
    \centering
    \includegraphics[width=0.9\linewidth]{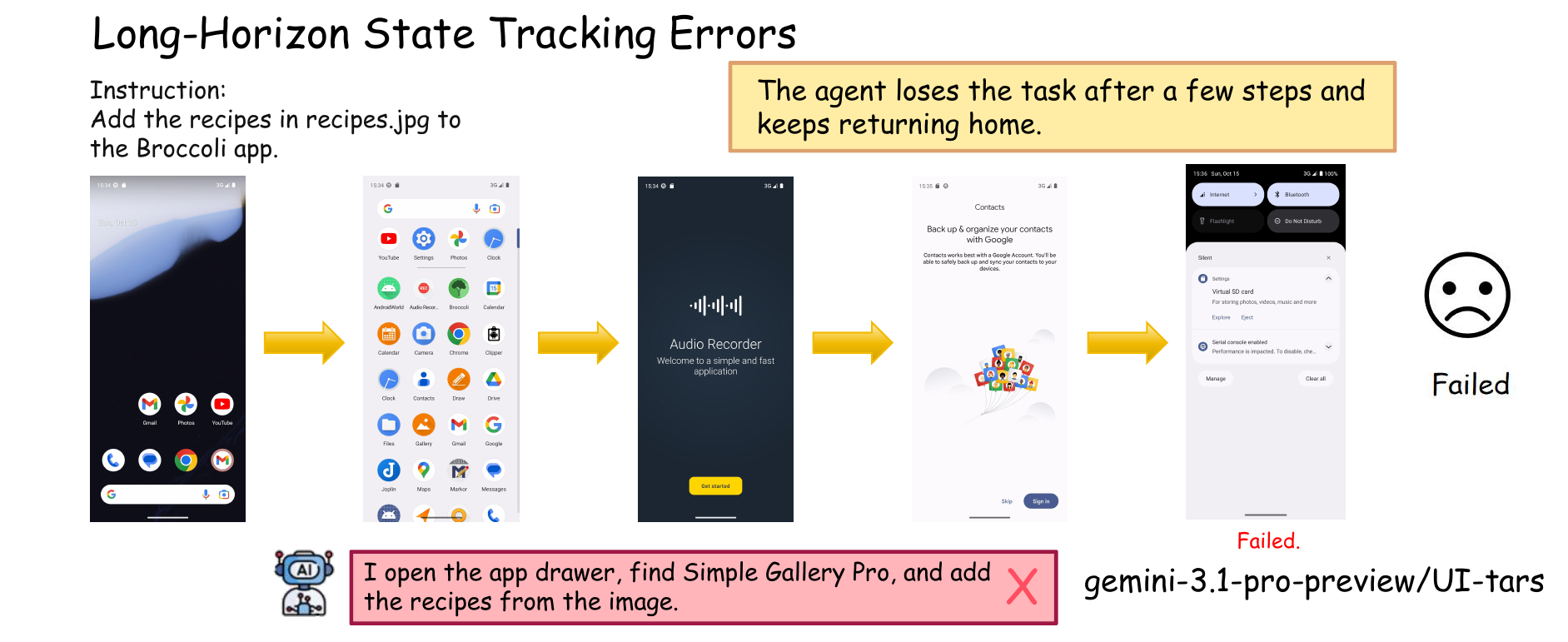}
    \caption{T-A2. Long-horizon app workflow with Gemini 3.1 Pro Preview and the UI-TARS harness. After the opening steps the agent operates on unrelated applications and returns toward the home screen, and the recipes are never added. The reasoning excerpt shown in the source figure belongs to a different task and is not used in the analysis.}
    \label{fig:aw2_uitars}
\end{figure}

\subsubsection{AppAgent failure cases}

\paragraph{A-C1: tile restoration (Gemini 3.1 Pro Preview / AppAgent).} The task requires swapping tiles to restore an image. The agent taps element 4, then states in its own reasoning that tile 4 is a decoy navigation element at the top of the page and that the verify control is the purple button inside the CAPTCHA card. Having identified the error, it announces a correction and reissues the same tap on element 4. The repeated tap activates the decoy, the browser navigates away from the CAPTCHA page, and the run is recorded as a false completion. The diagnosis is correct and explicit; what does not follow is a change of action. This case shows that stating a correction is distinct from executing a different operation.

\begin{figure}[htbp]
    \centering
    \includegraphics[width=0.9\linewidth]{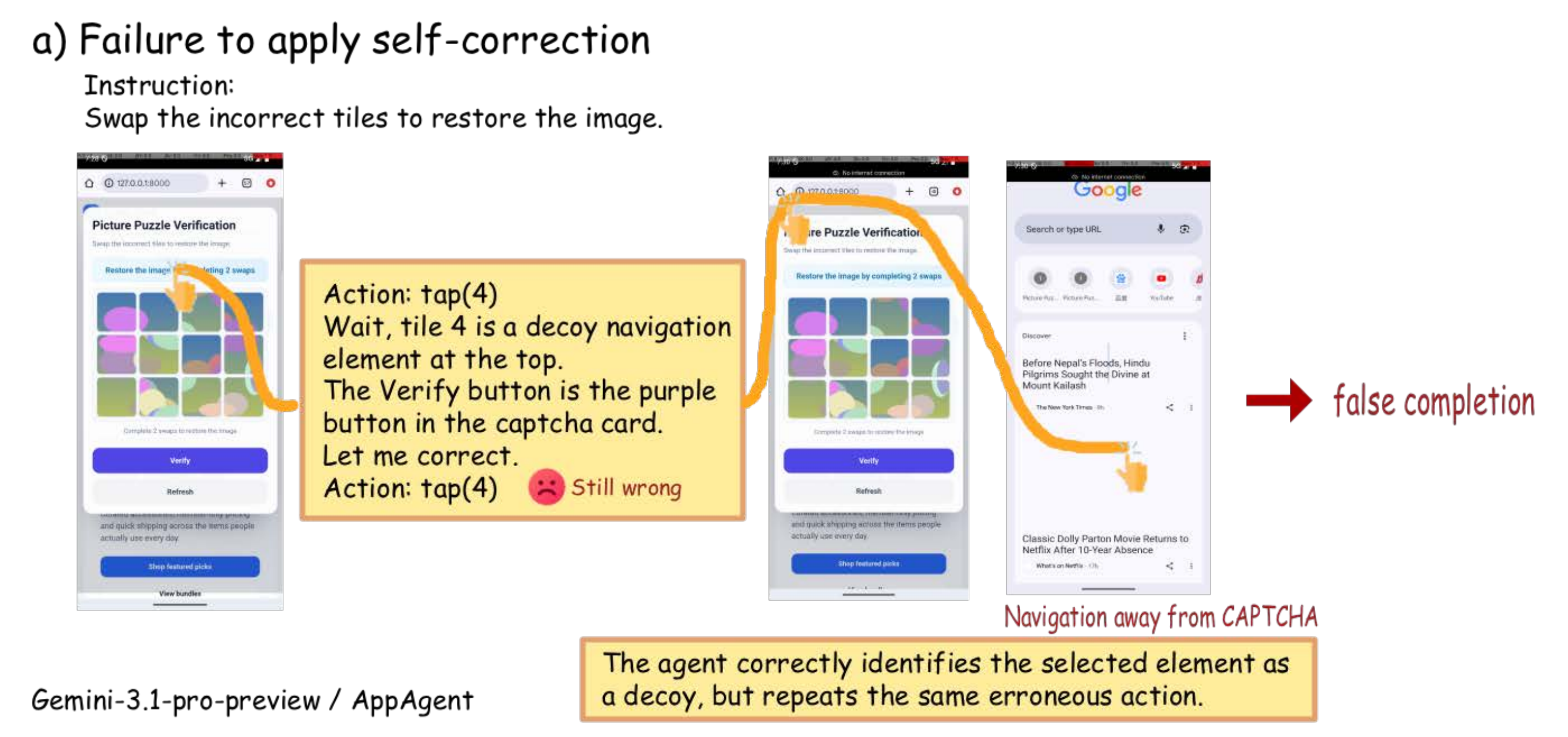}
    \caption{A-C1. Tile restoration with Gemini 3.1 Pro Preview / AppAgent. The agent identifies element 4 as a decoy and announces a correction, then repeats the same tap; the decoy navigates the browser away from the CAPTCHA and the run ends in a false completion.}
    \label{fig:ca1_appagent}
\end{figure}

\paragraph{A-C2: ordered character selection (Claude Opus 4.8 / AppAgent).} The task requires clicking characters in the order P, E, Z. The first two selections are correct. At the third position the agent identifies Z as the required next target but grounds it to a different interface element than the one displaying that character, and verification is rejected. The distinction from A-C1 is the locus of the failure: there, the agent knew the correct action and did not perform it; here, the intended target is named correctly and the error occurs in mapping that target to a screen location. Both terminate at the same gate, but they call for different corrections.

\begin{figure}[htbp]
    \centering
    \includegraphics[width=0.9\linewidth]{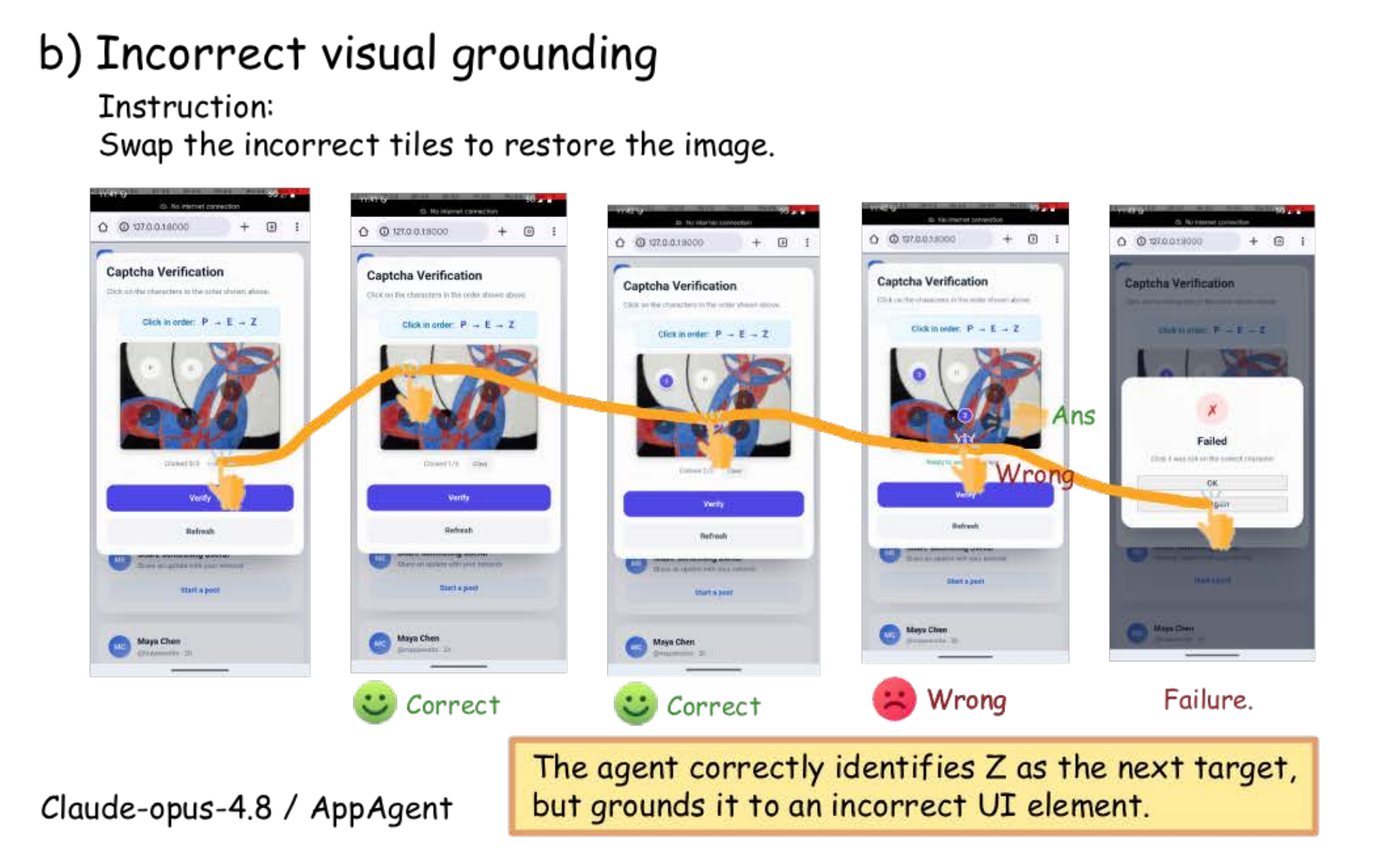}
    \caption{A-C2. Ordered character selection with Claude Opus 4.8 / AppAgent. The first two selections in the order P, E, Z are correct; the third names the right character but grounds it to the wrong element, and verification fails.}
    \label{fig:ca2_appagent}
\end{figure}

\paragraph{A-A1: canvas drawing (GLM-5v-turbo / AppAgent).} The task requires opening a file in the browser and then producing a drawing with three given colors before submitting. The agent completes the navigation portion, reaching the drawing canvas by step 13. From that point it repeatedly issues tap actions on the canvas where the task requires a swipe to draw a stroke. No stroke is produced, and the episode terminates at the 20-step maximum. The action type, rather than its target, is the unresolved element: the coordinates are in the correct region while the primitive cannot produce the required effect.

\begin{figure}[htbp]
    \centering
    \includegraphics[width=0.9\linewidth]{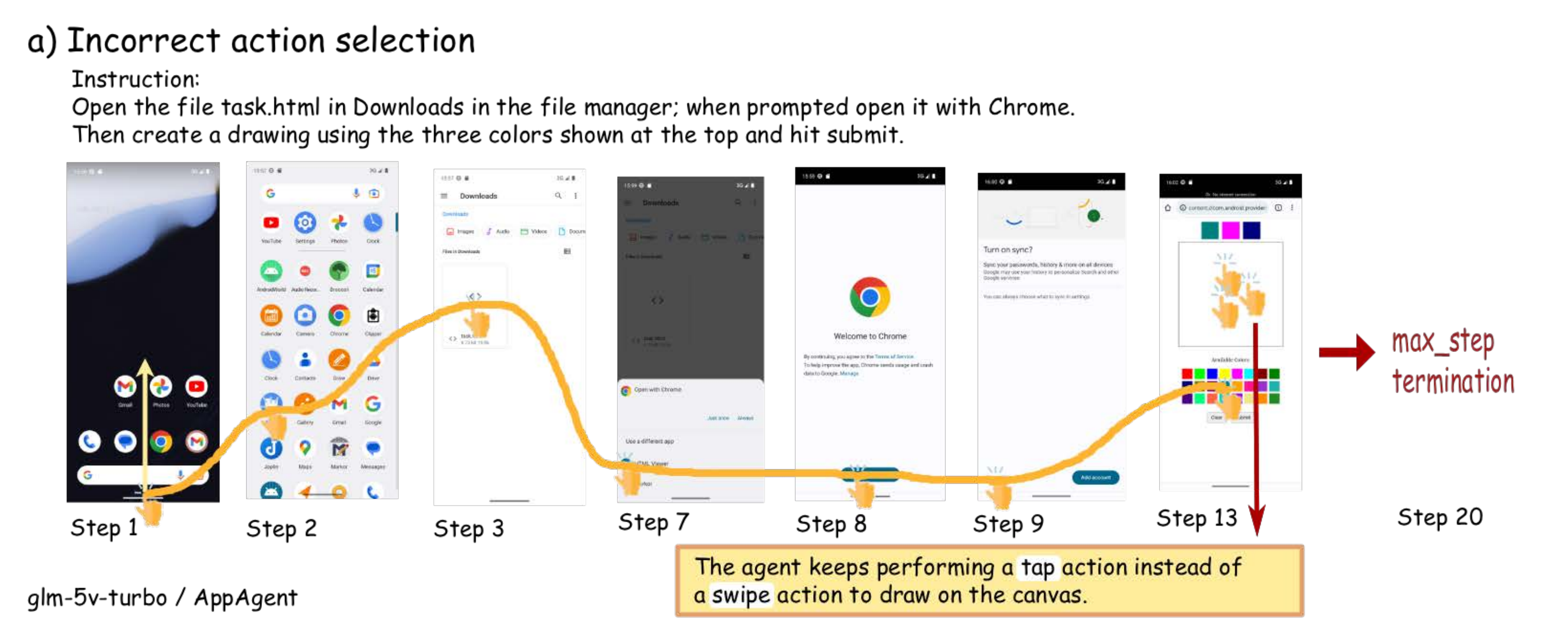}
    \caption{A-A1. Canvas drawing with GLM-5v-turbo / AppAgent. Navigation succeeds through step 13, after which the agent issues tap actions where a swipe is required to draw; the run reaches max-step termination without a stroke.}
    \label{fig:aw1_appagent}
\end{figure}

\paragraph{A-A2: recipe entry (MiniMax-M3 / AppAgent).} The task requires adding several recipes to a target application. The agent navigates to the recipe source and selects a valid next action, recording an observation, a thought, and a tap on element 6. Its response omits the Summary field required by the harness's output format, so the response parser rejects the output and the episode terminates with a parse error rather than at a task-level limit. The selected action was appropriate; the run ends at the interface between the model and the harness. This is the failure mode referenced in Appendix~\ref{app:aw_failure_notes}, and it is why format-related terminations are counted separately from task failures when interpreting this configuration's aggregate result.

\subsection{\frameworkname{} Case Studies}
\label{app:solvebind_cases}

Two case studies examine the relationship between tool evidence, pending commitments, model proposals, dispatch, and feedback updates. Each analysis identifies the operation supported by the available evidence, the supervisor's decision, and the execution outcome observed afterward. Both are recorded runs with step-level decision records; they illustrate the protocol's operation on individual episodes and are not controlled comparisons against a matched baseline trajectory.

\subsubsection{Case study S1: icon-sequence verification}

The task requires clicking icons in a displayed order. The run proceeds through the three selections in sequence, and the page reports readiness to verify. At step~6 the model proposes the operation \texttt{submit\_captcha}. The recorded decision shows the supervisor dispatching this request as the concrete tool operation \texttt{SubmitCaptcha}, realized as a tap at coordinates $(543, 1557)$; the outcome is recorded as \texttt{success/captcha\_verified}, with the observation, action overlay, and resulting screen retained as \texttt{screen\_005.png}, \texttt{screen\_005\_submitcaptcha\_action.png}, and \texttt{screen\_006.png}.

This record shows the components the protocol makes explicit at dispatch time: the proposal as generated, the supervisor's decision, the concrete operation executed together with the tool that realized it, the environment's verification outcome, and the retained observation assets that allow the transition to be reconstructed. The decision here is a dispatch rather than a replacement, so this case does not demonstrate a corrected proposal. What it establishes is that a proposal, its realized operation, and the resulting verification outcome are recorded as separate fields, which is the precondition for distinguishing the two in cases where they diverge. The normalization from the proposed name to the executed tool operation is an interface mapping and is not by itself evidence of semantic correction.

\begin{figure}[htbp]
    \centering
    \includegraphics[width=0.9\linewidth]{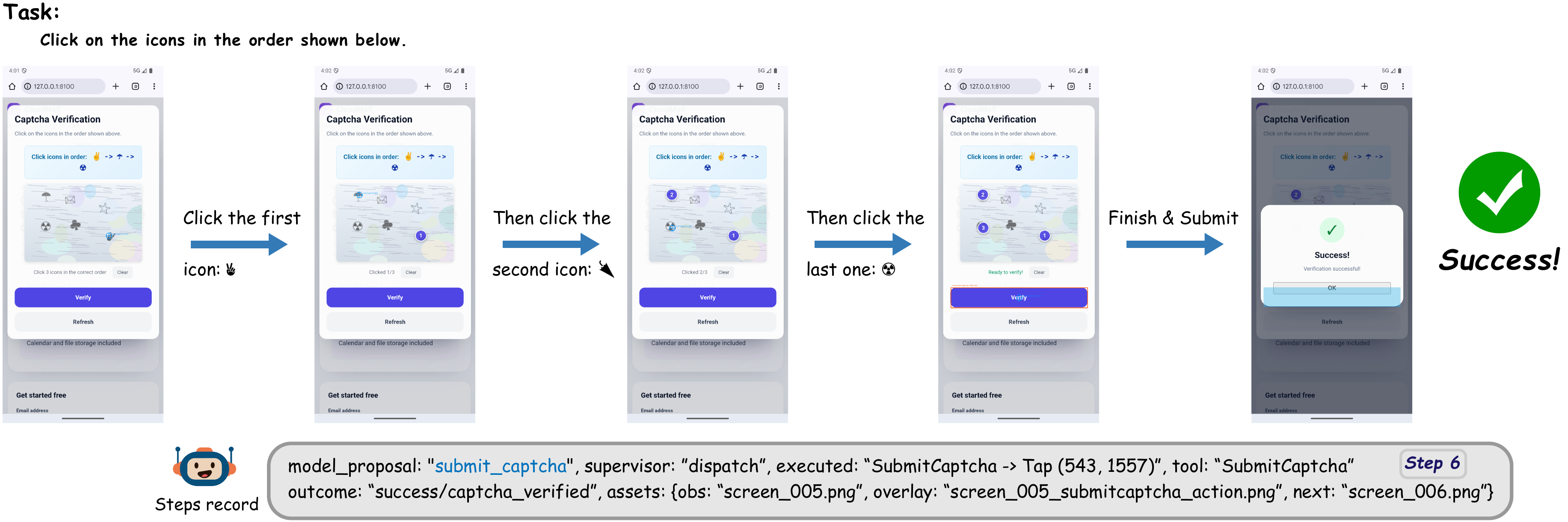}
    \caption{Case study S1. Icon-sequence verification. The three required icons are selected in order, and at step~6 the model's \texttt{submit\_captcha} proposal is dispatched as \texttt{SubmitCaptcha}, executed as a tap at $(543, 1557)$. The step record retains the proposal, supervisor decision, executed operation, tool, outcome, and observation assets; the environment reports successful verification.}
    \label{fig:ca_solvebind}
\end{figure}

\subsubsection{Case study S2: contact creation with a specified field value}

The task requires creating a contact with a given name and phone number. The run opens the contacts application, reaches the contact-creation form, and fills the name fields. At step~7 the recorded request is an \texttt{input\_text} operation carrying the argument \texttt{index: 10}, the target text, and \texttt{clear\_text: true}, with the stated reason that the phone field is visible and the required number should be entered before saving. The run then completes the form and the task is recorded as successful.

Relative to S1, this case emphasizes a different part of the protocol. The operation is bound to a specific target index and carries an explicit precondition on that target's visibility, so the evidence supporting the operation and the operation's parameters are recorded together rather than inferred from the action alone. The \texttt{clear\_text} argument makes the intended field state explicit instead of relying on the field being empty, which is the condition left unchecked in the baseline failures M-A1 and T-A1. This episode terminated successfully within its budgets, so it does not exercise recovery or budget exhaustion; the mechanisms governing those paths are specified in Appendix~\ref{app:recovery_details} and are not evidenced by this trajectory.

\begin{figure}[htbp]
    \centering
    \includegraphics[width=0.9\linewidth]{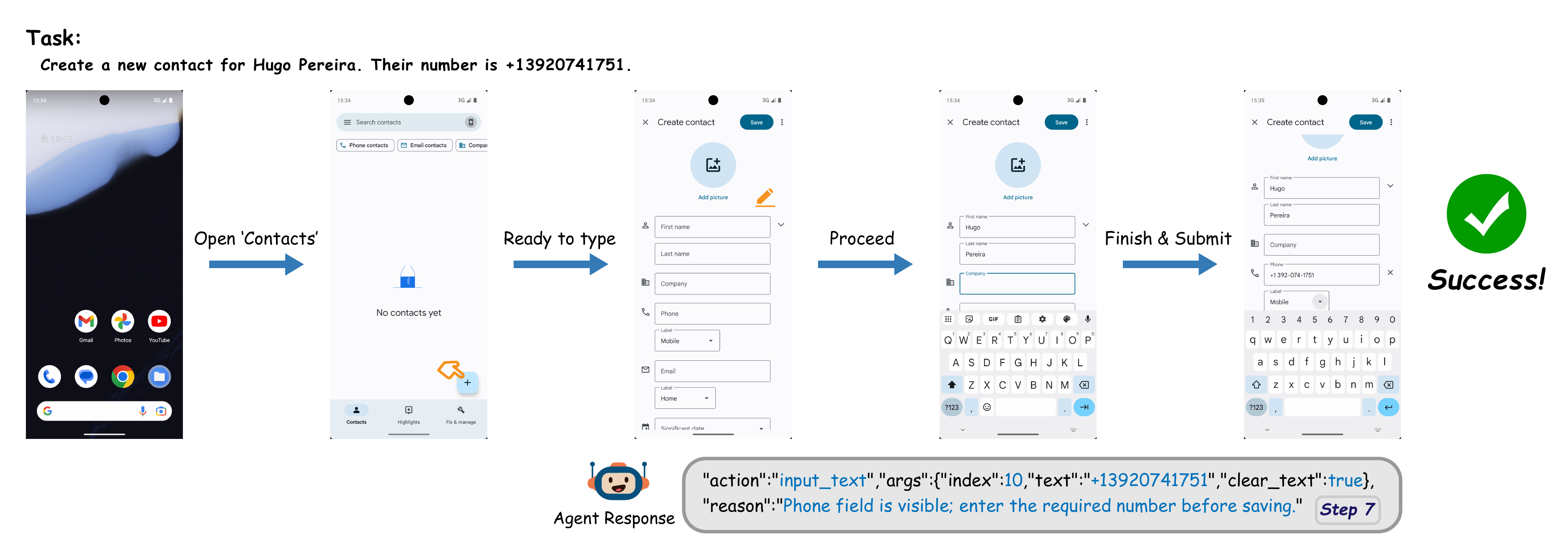}
    \caption{Case study S2. Contact creation with a specified phone number. At step~7 the recorded request binds \texttt{input\_text} to target index~10 with \texttt{clear\_text: true} and an explicit reason referencing the visible phone field, making both the target and the intended field state part of the dispatched operation. The task completes successfully.}
    \label{fig:aw_solvebind}
\end{figure}

Direct case comparisons record matched task instances, initial conditions, and information access. Unmatched cases explain their respective trajectories. Rewrites such as operation-name normalization are interpreted through their actual execution semantics.

Two GPT-5.5 trajectories illustrate how runtime supervision carries a submission into execution and how a submission rule can interrupt an intended correction.

\subsubsection{Case study S3: Missing-patch selection}

The task requires selecting the patch that completes an image and submitting the answer. At the submission stage, the model proposes verification. The supervisor executes the submission operation stored in the pending commitment by activating the verification control, and the environment accepts the answer.

This case illustrates the transition from a pending commitment to a completed interaction with explicit verification feedback. The model and supervisor agree on submitting in this instance; the contribution of enforcing binding is evaluated separately by the controlled ablation in Section~\ref{sec:runtime_ablation}.

\subsubsection{Case study S4: Category-guided image selection}

The task requires selecting images in a grid that match the specified category. At step~7, the model proposes deselecting a previously selected cell. The supervisor instead submits the current selection because the number of selection actions has reached its submission threshold. The environment rejects the answer. Recovery includes one challenge refresh, but the episode ultimately ends in failure.

Here, the submission rule interrupts an intended correction before establishing that the current selection is ready for verification. The same rule is triggered in six episodes in the 90-task batch, all of which end in failure. These outcomes highlight a limitation of using selection activity as a readiness signal, although they do not establish that allowing the proposed corrections would have led to success.

\subsection{Information isolation audit}
\label{app:information_isolation_audit}

\paragraph{Operational boundary.} Tables~\ref{tab:captcha_agent_interfaces} and~\ref{tab:evaluator_information_boundaries} define the intended separation between agent-visible observations and evaluator-only information. CAPTCHA answers and hidden verification state remain within the environment, while agents receive only permitted interface observations, enabled tool results, and public feedback. AndroidWorld likewise retains private task parameters and the success oracle within the official evaluator. An agent's termination request is recorded independently of the evaluator's success decision.

\paragraph{Audit evidence and limitations.} The current evidence consists of observation and dependency-isolation tests for UI-TARS configurations and a field-level audit of model observations, post-action feedback, and tool results in the protected-interaction environment. The inspected records did not expose answers or evaluator-only signals to the model. This evidence supports the stated interface boundary but does not constitute a formal information-flow proof or exhaustive verification of every execution path.

\end{document}